\documentclass[12pt,authoryear]{elsarticle/elsarticle}

\usepackage{amssymb,amsmath,graphicx}
\usepackage{url}
\usepackage{enumerate}
\usepackage{multirow}
\usepackage{tipa}
\usepackage{tabularx}
\usepackage{tikz}
\usepackage{tikzscale}
\usepackage{dblfloatfix}
\usepackage{attachfile2}
\usepackage{tabu}
\usepackage{multicol}
\usepackage{lipsum}
\usepackage{color}
\usepackage{textcomp}
\usepackage{caption}
\usepackage{subcaption}

\usepackage{pgfplots}

\usetikzlibrary{fit}
\tikzset{%
  highlight/.style={rectangle,rounded corners,fill=red!15,draw,fill opacity=0.5,thick,inner sep=0pt}
}
\newcommand{\tikzmark}[2]{\tikz[overlay,remember
  picture,baseline=(#1.base)] \node (#1) {#2};}
\newcommand{\Highlight}[1][submatrix]{%
    \tikz[overlay,remember picture]{
    \node[highlight,fit=(left.north west) (right.south east)] (#1) {};}
}

\newcommand{\ph}{\textipa}
\newcommand{\vt}{\rotatebox{90}}
\newcolumntype{"}{@{\hskip\tabcolsep\vrule width 1pt\hskip\tabcolsep}}

\newcommand{\rv}{\color{black}}

\renewcommand{\vec}[1]{\mathbf{#1}}

\begin{document}

\begin{frontmatter}



\title{Speech vocoding for laboratory phonology}


\author[label1]{Milos Cernak\corref{cor1}}
\ead{milos.cernak@idiap.ch}
\author[label2]{Stefan Benus}
\author[label1]{Alexandros Lazaridis}

\address[label1]{Idiap Research Institute, Martigny, Switzerland}
\address[label2]{Constantine the Philosopher University in Nitra, Slovakia and Institute of Informatics, Slovak Academy of Sciences, Bratislava, Slovakia}
\cortext[cor1]{Corresponding author}

\begin{abstract}
Using phonological speech vocoding, we propose a platform for
exploring relations between phonology and speech processing, and in
broader terms, for exploring relations between the abstract and
physical structures of a speech signal. Our goal is to make a step
towards bridging phonology and speech processing and to contribute to
the program of Laboratory Phonology.

We show three application examples for laboratory phonology:
compositional phonological speech modelling, a comparison of
phonological systems and an experimental phonological parametric
text-to-speech (TTS) system. The featural representations of the
following three phonological systems are considered in this work: (i)
Government Phonology (GP), (ii) the Sound Pattern of English (SPE),
and (iii) the extended SPE (eSPE).
Comparing GP- and eSPE-based vocoded speech, we conclude that the
latter achieves slightly better results than the former. However, GP
-- the most compact phonological speech representation -- performs
comparably to the systems with a higher number of phonological
features. The parametric TTS based on phonological speech
representation, and trained from an unlabelled audiobook in an
unsupervised manner, achieves intelligibility of 85\% of the
state-of-the-art parametric speech synthesis.

We envision that the presented approach paves the way for researchers
in both fields to form meaningful hypotheses that are explicitly
testable using the concepts developed and exemplified in this
paper. On the one hand, laboratory phonologists might test the applied
concepts of their theoretical models, and on the other hand, the
speech processing community may utilize the concepts developed for the
theoretical phonological models for improvements of the current
state-of-the-art applications.

\end{abstract}


\end{frontmatter}


\section{Introduction}




Speech is a domain exemplifying the dichotomy between
the continuous and discrete aspects of human behaviour. 
On the one hand, the articulatory activity and the resulting
{\rv acoustic} speech signal are continuously varying. On the other
hand, for speech communication to convey meaning, this continuous
signal must be, at the same time, perceivable as {\rv
  contrastive}. Traditionally, these two aspects have been studied
within phonetics and phonology respectively. Following significant
successes of this dichotomous approach, for example in speech
synthesis and recognition, recent decades have witnessed a lot of
progress in understanding and formal modelling of the relationship
between these two aspects, e.g. the program of Laboratory Phonology
\citep{Pierrehumbert00} or the renewed interest in the approaches
based on Analysis by Synthesis \citep{Hirst11,Bever10}. The goal of
this paper is to follow these developments by proposing a platform for
exploring relations between the mental (abstract) and physical
structures of the speech signal. In this, we aim at mutual
cross-fertilisation between phonology, as a quest for understanding
and modelling of cognitive abilities that underlie systematic patterns
in our speech, and speech processing, as a quest for natural, robust,
and reliable automatic systems for synthesising and recognising
speech.

As a first step in this direction we examine a cascaded speech
analysis and synthesis approach (known also as vocoding) based on
phonological representations and how this might inform both quests
mentioned above. In parametric vocoding speech segments of different
time-domain granularity, ranging from speech frames, e.g. in the 
formant~\citep{Holmes1973}, or articulatory~\citep{Goodyear1996,
  Laprie2013} domains, to phones \citep{Tokuda98,Lee2001}, and syllables
\citep{Cernocky98}, are used in sequential processing.
In addition to these segments, phonological representations have also been shown
to be useful for  speech processing e.g. by~\cite{King00}. In our
work, we explore a direct link between phonological features and their
engineered acoustic realizations. In other words, we believe that
abstract phonological sub-segmental, segmental, and suprasegmental
structures may be related to the physical speech signal through a
speech engineering approach, and that this relationship is informative
for both phonology and speech processing.

The motivation for this approach is two-fold. Firstly, phonological
representations (together with grammar) create a formal
model whose overall goal is to capture the core properties of the
cognitive system underlying speech production and perception.
This model, linking subsegmental, segmental, and suprasegmental
phonological features of speech, finds independent support in the
correspondence between a) the brain-generated cortical oscillations in
the `delta' (1-3 Hz), `theta' (4-7 Hz), and faster `gamma' ranges
(25-40 Hz), and b) the temporal scales for the domains of prosodic
phrases, syllables, and certain phonetic features respectively. In
this sense, we may consider phonological representations
embodied~\citep{Giraud12}.
Hence, speech processing utilizing such a system might
lead to a biologically sensible and empirically testable computational
model of speech.

Secondly, phonological representations are inherently
multilingual~\citep{Siniscalchi2012}. This in turn has an attractive
advantage in the context of multilingual speech processing in
lessening the reliance on purely phonetic decisions. The independence
of the phonological representations from a particular language on the
one hand and the availability of language specific mapping between these
representations and the acoustic signal through speech processing
methods on the other hand, offer (we hope) a path towards a context-based
interpretation of the phonological representation that is grounded in
phonetic substance but at the same time abstract enough to allow for a
more streamlined approach to multilingual speech processing.

In this work, we propose to use the phonological
vocoding of~\cite{Cernak15} and other advances of speech processing for testing certain aspects of phonological theories. 
We consider the following phonological systems
in this work:
\begin{itemize}
  \item The Government Phonology (GP) features \citep{Harris95}
  describing sounds by fusing and splitting of 11 primes.
  \item The Sound Pattern of English (SPE) system with 13 features
  established from natural (articulatory)
  features~\citep{chomsky68sound}.
  \item The extended SPE system (eSPE)
  ~\citep{Yu2012,Siniscalchi2012} consisting of 21
  phonological features.
\end{itemize}
Having trained phonological vocoders for the three phonological models
of sound representation, we describe several application examples
combining speech processing techniques and phonological representations.
Our primary goal is to demonstrate the usefulness of
the analysis by synthesis approach by showing that (i) the vocoder can
generate acoustic 
realizations of phonological features used by compositional speech
modelling, (ii) speech sounds (both individual sounds not seen in
training and intelligible continuous speech) can be generated from the
phonological speech representation, and (iii) the testing of
hypotheses relating phonetics and phonology is possible; we test the
hypothesis that the best phonological speech representation achieves
the best quality vocoded speech, by evaluating the phonological
features in both  directions, recognition and synthesis,
simultaneously. Additionally, we compare the segmental properties of
the phonological systems, and describe results and
advantages of experimental phonological parametric TTS
system. To preview, conventional parametric TTS can be described as phonetic, i.e., phonetic and other 
linguistic and prosodic information is used in the input labels. 
On the other hand, our proposed system can be described as phonological, i.e., 
only information based on phonological representation is used as input. 
This allows, for example, for generation of any new speech sounds.

The structure of the paper is as follows: the phonological
representations used in this work are introduced in
Section~\ref{sec:phonology}. Section~\ref{sec:phonovoc} introduces
speech vocoding based on phonological speech
representation. Section~\ref{sec:experiments} describes the
experimental setup. The application examples of the proposed platform
(along with the experiments and results) are shown in
Section~\ref{sec:examples}. Finally the conclusions follow in
Section~\ref{sec:conclusions}.

\section{Phonological systems}
\label{sec:phonology}


Phonology is construed in this work as a formal model that
  represents
cognitive (subconscious) knowledge of native speakers regarding the
systematic sound patterns of a language. The two traditional
components of such models are i) system primitives, that is, the units
of representation for cognitively relevant objects such as sounds or
syllables, and ii) a set of permissible operations over these
representations that is able to generate the observed
patterns. Naturally, these two facets are closely linked and
inter-dependent. In this paper we focus on the theory of representation.

\subsection{Segmental representations}

The minimal unit of meaning contrast, i.e. cognitive relevance, is
assumed to be the phoneme. In the tradition of~\cite{Jakobson56}
and~\cite{chomsky68sound}, phonemes are assumed to consist of feature
bundles. In the former model of~\cite{Jakobson56}, 12 basic perceptual-acoustic domains
(e.g. acute-grave, or compact-diffuse) define the space for
characterising all the phonemes. The model uses polar opposites for
these 12 continua, which are necessarily relational, and thus
language-specific. Hence, a vowel characterised as, e.g. ‘grave’ in
one language might be phonetically different from the same grave vowel
in another language since their grave quality might be at a different
point of the acute-grave continuum. The latter system
of~\cite{chomsky68sound}, known also as SPE, differed 
from the former system of~\cite{Jakobson56} in two fundamental aspects
relevant for this 
paper. First, it took the articulatory production mechanism as the
underlying principle of phoneme organisation; hence, in their 13 basic
binary features, we talk about the position (or activity) of the
active articulators rather than percepts they
create. Second, SPE assumed that the flat, unstructured binary feature
specifications are language independent and characterise the set of
possible phonemes in languages of the world.


The developments of phonological theory after SPE focused on both the
theory of representations as well as the operations. The most relevant
for this paper are proposals for establishing the non-linear nature of
phonological representations, i.e., the fact that individual features
are not strictly linked to the linear sequence of sounds but may span
greater domains or occupy independent non-overlapping tiers. These proposals started
with the representation of
lexical tones \citep{Leben73phd,Goldsmith76phd}, continued with the
featural geometry approach \citep{Sagey86phd,Clements95} and received
novel formal treatments in the theories of Dependency and
Government Phonology (GP), e.g. \cite{Kaye90}, \cite{Harris94}. These
latter approaches posit the so called primes, or basic elements, that are
monovalent (c.f. binary SPE features). For example, there are four
basic resonance primes commonly labelled as A, U, I, and @; the first
three denoting the peripheral vowel qualities \ph{[a]}, \ph{[u]} and \ph{[i]}
respectively, the last one describing the most central vowel quality of
schwa. English \ph{[i:]} would correspond to the I prime while \ph{[e]}
results in fusing the I and A primes. In addition to these `vocalic'
primes, GP proposes the `consonantal' primes ?, h, H, N, denoting
closure, friction, voicelessness and nasality respectively. To account
for the observed variability in inventories (e.g. English has 20 contrastive vowels
and thus 20 different phonetic representations for them), phonetic
qualities, and types of phonological behaviour, 
the phoneme representations based on simple primes become
insufficient. The solution, given the name of the framework, is that
some elements can optionally be heads
and govern the presence or realisation of other (dependent) elements.

These developments established the relevance of the internal structure
of phonological primitives and their hierarchical
nature. Furthermore, the non-linear nature of phonological
representations became the
accepted norm for formal theories of phonology. Importantly, while the
SPE-style features were assumed to require the full interpretation of
all features for a surface phonetic realisation of a phoneme, the
primes of GP are assumed to be interpretable alone despite their
sub-phonemic nature. \cite{Harris95} call this GP assumption the
autonomous interpretation hypothesis.

This hypothesis, and its testing, is at the core of our approach. One
of the goals of this work is to provide an interpretation for
sub-phonemic representational components of phonology that is grounded
in the acoustic signal.

\subsection{Representing CMUbet with SPE and GP}

To characterise the phoneme inventory of American English in the SPE
and the GP frameworks, we have adopted the approach of~\cite{King00}
with some modifications. We use the reduced set of 39 phonemes
in the CMUbet
system\footnote{\url{http://www.speech.cs.cmu.edu/cgi-bin/cmudict}}. The
major difference regarding the SPE-style representations is our 
addition of a [rising] feature used to differentiate diphthongs from
monophthongs. In the original SPE framework, this difference was
treated with the [long] feature and the surface representation of
diphthongs was derived from long monophthongs through a rule. Given
the absence of the `rule module' in our approach to synthesis, we
opted for a unified feature specification of the full vowel inventory
of American English using the added feature. This allowed for
diphthongs to form a natural class with vowels rather than with glides
\ph{[j, w]} as in~\cite{King00}. Additional minor adjustments 
 assured the uniqueness of a feature matrix for each
phoneme. The full specification of all 39 phonemes with 14 binary
features is shown in Table~\ref{tab:SPEatts}.

The set of GP-style specification for English phonemes can be seen in
Table~\ref{tab:GPatts}. Again, we followed King \& Taylor,
mostly in formalizing headedness with pseudo-features, which allows
for GP phoneme specifications that are comparable to SPE.
Additionally, the vowel specifications differ somewhat from King \& Taylor
stemming from our effort to approximate the phonetic characteristics
of the vowels, and the differences among them, as closely as
possible. For example, if the back lax \ph{[U]} is specified with E,
the same was employed for the front lax \ph{[I]}, or, the front quality
of \ph{[\ae]} was formalized with A heading I compared to King \&
Taylor's formalism with only non-headed A.





\section{Phonological vocoding}
\label{sec:phonovoc}
Both SPE and GP phonological systems represent a phone by a
combination of phonological features. For example, a consonant 
\ph[j] is articulated using the mediodorsal part of the tongue
[+Dorsal], in the mediopalatal part of the vocal tract
[+High], generated with simultaneous vocal fold vibration
[+Voiced]. These three features then comprise the phonological
representation for \ph[j] in the SPE system.

\begin{figure}[!t]
\centering
\resizebox{4.5in}{!}{%
\begin{tikzpicture}[font=\scriptsize]
  \draw[rounded corners, thick] (0,1.25) rectangle (1,1.75); \node at
  (0.5,1.5) {Speech};

  \draw[rounded corners, thick] (1.75,1) rectangle (3,2);
  \node[align=center] at (2.4,1.5) {Speech\\analysis};
  \draw[thick, ->] (1,1.5) -- (1.75,1.5);
  \draw[thick, ->] (3,1.5) -- (4,1.5);
  \node[align=center] at (5,3.5) {\textbf{Phonological analysers}};
  \draw[rounded corners, thick, fill=lightgray] (4,2.5) rectangle (6,3);
  \node[align=center] at (5,2.75) {Segmental};

  \draw[thick, <->] (4,2.75) -- (3.5,2.75) -- (3.5,0.25) -- (4,0.25);
  \draw[rounded corners, thick, fill=lightgray] (4,1) rectangle (6,2);
  \node[align=center] at (5,1.5) {Supra-\\segmental};
  \draw[thick] (6,2.75) -- (6.5,2.75) -- (6.5,0.25) -- (6,0.25);

  \draw[rounded corners, thick, fill=lightgray] (4,0) rectangle (6,0.5);
  \node[align=center] at (5,0.25) {Silence};

  \draw[thick, ->] (6,1.5) -- (7.5,1.5); \node[above] at (7,1.5)
       {$\vec{z}_n$};

  \draw[rounded corners, thick, fill=lightgray] (7.5,1) rectangle (9.5,2);
  \node[align=center] at (8.5,1.5) {\textbf{Phonological}\\\textbf{synthesizer}};

  \draw[thick, ->] (9.5,1.8) -- (11,1.8); \node[above] at (10.25,1.8)
       {LSPs};

  \draw[thick, fill=black] (10,1.5) circle[radius=0.04];
  \draw[thick, fill=black] (10.25,1.5) circle[radius=0.04];
  \draw[thick, fill=black] (10.5,1.5) circle[radius=0.04];

  \draw[thick, ->] (9.5,1.2) -- (11,1.2); \node[align=center] at (10.25,0.8)
       {Glottal\\signal};

  \draw[rounded corners, thick] (11,1) rectangle (13,2);
  \node[align=center] at (12,1.5) {LPC\\re-synthesis};

\end{tikzpicture}
}%
\caption{The process of the phonological vocoding. The speech signal
  is internally represented by phonological posterior probabilities
  $\vec{z}_n$, consisting of $K$ phonological posteriors per $n$-th
  frame). The DNN-based synthesizer generates speech parameters --
  line spectral pairs (LSPs) and source parameters for LPC speech
  re-synthesis.}
\label{fig:phonovocTRN}
\end{figure}
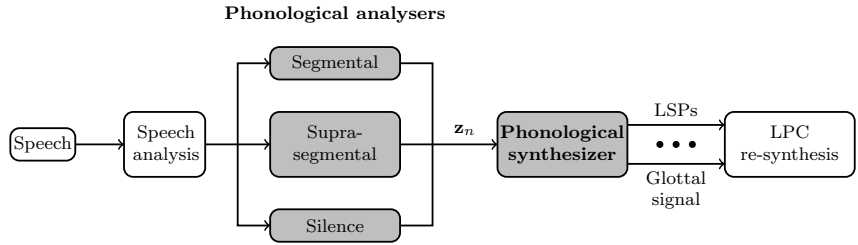

\cite{Cernak15} have recently designed a phonological vocoder for a
low bit rate parametric speech coding as a cascaded artificial neural
network composed of speech analyser and synthesizer that use a shared
phonological speech representation. Figure~\ref{fig:phonovocTRN} shows
a sequential processing of the vocoding, briefly introduced in the
following text.

\subsection{Phonological analysis}

Phonological analysis starts with speech analysis that turns speech
samples into a sequence of acoustic feature observations
$X=\{\vec{x}_1,\ldots,\vec{x}_n,\ldots,\vec{x}_N\}$ where $N$ denotes
the number of frames in the speech signal. Conventional cepstral
coefficients can be used in this speech analysis step. Then, a bank of
phonological analysers realised as neural network 
classifiers converts the acoustic feature observation sequence $X$
into a sequence of vectors
$Z=\{\vec{z}_1,\ldots,\vec{z}_n,\ldots,\vec{z}_N\}$ where a vector
$\vec{z}_n=[z_n^1,\ldots,z_n^k,\ldots,z_n^K]^T$ consists of $K$
phonological feature posterior probabilities, called hereinafter
the phonological posteriors.
The posteriors $z_n^k$ are probabilities that the $k$-th feature
occurs (versus does not occur), and they compose the
information which is transferred to the synthesizer's side in order to
synthesize speech as described in the next subsection. These
posteriors can be optionally pruned (compressed) if there is need for
reducing the amount of transferred information (i.e. reducing the
bit rate). For example, the binary nature of the phonological features
considered by~\cite{Cernak15} allowed for using binary values of the
phonological features instead of continuous values, which resulted
only in minimal perceptual degradation of speech quality.

There are two main groups of the phonological features:
\begin{itemize}
  \item segmental: phonological features that define the phonetic
    surface of individual sounds,
  \item supra-segmental: phonological features at the timescales of
    syllables and feet (including a single stressed syllable and one
    or more unstressed ones.
\end{itemize}
In this work, we focus on the segmental phonological features and leave the supra-segmental for future research.


\subsection{Phonological synthesis}
\label{sec:resynthesis}

The phonological synthesizer is based on a Deep Neural Network (DNN)
that learns the highly-complex regression problem of mapping
posteriors $z_n^k$ to the speech parameters. More specifically, it
consists of two computational steps. The first step is a DNN forward
pass that generates the speech parameters, and the second one is a
conversion of the speech parameters into the speech samples. 
The second stage of synthesis is based on an open-source LPC
re-synthesis with minimum-phase complex cepstrum glottal model
estimation~\citep{Phil15}. The modelled speech parameters are thus:
\begin{itemize} 
  \item $\vec{p}_n$: static Line Spectral Pairs (LSPs) of 24th order
    plus gain, 
  \item $\log(\vec{r}_n)$: a Harmonic-To-Noise (HNR) ratio,
  \item and $\vec{t}_n$, $\log(\vec{m}_n)$: two glottal model
    parameters -- {\rv angle and magnitude of a glottal pole,
      respectively.}
\end{itemize}

The generated speech parameter vectors -- $\vec{p}_n$,
$\vec{t}_n$, $\log(\vec{r}_n)$ and $\log(\vec{m}_n)$ for the $n$-th
frame -- from the first computational step are smoothed using dynamic
features and pre-computed (global) variances~\citep{Tokuda95}, and
formant enhancement~\citep{Ling2006} is performed to compensate for
over-smoothing of the formant frequencies. The speech samples are
finally generated frame by frame followed with overlap-and-add.
LPC re-synthesis can be done either with synthesised or original pitch
features.

\section{Experimental setup for laboratory phonology}
\label{sec:experiments}

In this section, the experimental protocol of the phonological
analysis and synthesis is described. In Table~\ref{tab:data}, the data
used in these experiments 
are presented. The analyser is trained on the Wall Street Journal
(WSJ0 and WSJ1) continuous speech recognition corpora \citep{WSJDB}.
The \textit{si\_tr\_s\_284} set of 37,514 utterances was used, split
into 90\% training and 10\%  cross-validation sets. The synthesizer is
trained and evaluated on an English audiobook ``Anna Karenina'' of
Leo
Tolstoy\footnote{\url{https://librivox.org/anna-karenina-by-leo-tolstoy-2/}},
that is around 36 hours long. Recordings were organised into 238
sections, and we used sections 1--209 as a training set, 210--230 as a
development set and 231--238 as a testing set. The development and
testing sets were 3 hours and 1 hour long, respectively.

\begin{table} [th]
  \caption{\label{tab:data} {\it Data used for training the
  phonological analysis and synthesis.}}
\vspace{2mm}
\centerline{
\begin{tabu}{|c|c|c|c|}
\hline 
Use of data & Database & Set/Section & Size \\
\hline  \hline
Analyser (train. set) & WSJ & si\_tr\_s\_284 & 33.765 (utts) \\
Analyser (cross-val. set) & WSJ & si\_tr\_s\_284 & 3.749 (utts) \\
Synthesizer (train. set) & Tolstoy & 1-209 & 36 (hours) \\
Synthesizer (dev. set) & Tolstoy & 210-230 & 3 (hours) \\
Synthesizer (test set) & Tolstoy & 231-238 & 1 (hour) \\
\hline
\end{tabu}}
\end{table}

\subsection{Phonological analysis}

The analyser is based on a bank of phonological analysers
realised as neural network classifiers -- multilayer perceptrons
(MLPs) -- that estimate phonological posteriors. Each MLP
is designed to classify a binary phonological feature. 

First, we trained a phoneme-based automatic speech recognition system
using Perceptual Linear Prediction (PLP) features. The phoneme set
comprising of 40 phonemes (including ``sil'', representing silence)
was defined by the CMU pronunciation dictionary. The three-state,
cross-word triphone models were trained with the HTS
variant~\citep{Zen:HTS} of the HTK toolkit on the 90\% subset of the 
\textit{si\_tr\_s\_284} set. The remaining 10\% subset was used for
cross-validation. We tied triphone models with decision tree state
clustering based on the minimum description length (MDL)
criterion~\citep{shinoda:mdl}. The MDL criterion allows an
unsupervised determination of the number of states. In this study, we
obtained 12,685 states and modeled each state with a Gaussian mixture
model (GMM) consisting of 16 Gaussians.


\subsubsection{Alignment}

A bootstrapping phoneme alignment was obtained using forced
alignment with cross-word triphones. The bootstrapping alignment was
used for the training of the bootstrapping MLP, using as the input 39
order PLP features with the temporal context of 9 successive frames,
and a softmax output function. The architecture of the MLP, 3-hidden
layers $351 \times 2000 \times 500 \times 2000 \times 40$ (input
$351=39 \times 9$ and output 40 is the number of phonemes), was
determined  empirically. Using a hybrid speech decoder fed with the
phoneme posteriors, the re-alignment was performed. After two
iterations of the MLP training and re-alignment, the best phoneme
alignment of the speech data was obtained. This re-alignment increased
the cross-validation accuracy of the MLP training from 77.54\% to
82.36\%.

\subsubsection{Training of the bank of phonological analysers}

The representation given in \ref{sec:spefeatures} was used to
map the phonemes of the best alignment to phonological features 
for the training of the
analysers. $K$ analysers were trained with the frame alignment having
two output labels, the $k$-th phonological feature occurs for the
aligned phoneme or not. The analysis MLPs were finally trained with
the same input features as used for the alignment MLP, and the same
network architecture except for 2 output units instead of 40. The
output that encodes the phonological feature occurrence is used as the
vector of phonological posteriors $\vec{z}_n$.

Tables~\ref{tab:accuracies}, \ref{tab:SPEaccuracies}, and
\ref{tab:eSPEaccuracies} show classification accuracies at frame
level of the GP, SPE and eSPE MLPs for the ``feature occurs'' output,
respectively. The accuracies are reported for the training and
cross-validation sets (see Table~\ref{tab:data} for data sets
definition). The analysers performances are high, with an average
cross-validation training accuracy of 95.5\%, 95.6\% and 96.3\%,
respectively. It is interesting that as the number of phonological
features increases, the classification accuracy also increases. The
reason might be that a more precise representation of the phonological
features is used (the maps given in \ref{sec:spefeatures}).

\begin{table} [t]
  \caption{\label{tab:accuracies} {\it Classification accuracies (\%)
      of the GP prime analysers at frame level on cross-val. set
      of \textit{si\_tr\_s\_284} set.}}
\vspace{2mm}
\centerline{
\begin{tabu}{|c|c|c|[0.8pt]c|c|c|}
\hline
Primes & \multicolumn{2}{c|[0.8pt]}{Accuracy (\%)} & Primes &
\multicolumn{2}{c|}{Accuracy (\%)} \\
\cline{2-3} \cline{5-6}
 & train & cv & & train & cv \\
\hline \hline
A & 92.3 & 91.7 & a & 97.9 & 97.6 \\
I & 94.9 & 94.6 & i & 96.1 & 96.4 \\
U & 94.4 & 94.1 & u & 97.5 & 97.7 \\
E & 92.6 & 91.9 & H & 95.4 & 95.0 \\
S & 95.2 & 94.7 & N & 98.2 & 98.1 \\
h & 95.9 & 95.4 & sil & 99.0 & 98.9  \\
\hline
\end{tabu}}
\end{table}

\begin{table} [t]
  \caption{\label{tab:SPEaccuracies} {\it Classification accuracies
      (\%) of the SPE analysers at frame level on cross-val. set
      of \textit{si\_tr\_s\_284} set.}} 
\vspace{2mm}
\centerline{
\begin{tabu}{|l|c|c|[0.8pt]l|c|c|}
\hline
Phonological & \multicolumn{2}{c|[0.8pt]}{Accuracy (\%)} & Phonological &
\multicolumn{2}{c|}{Accuracy (\%)} \\
\cline{2-3} \cline{5-6}
features & train & cv & features & train & cv \\
\hline \hline
vocalic & 96.0 & 95.8 & round & 97.8 & 97.7 \\
consonantal & 94.5 & 93.9 & tense & 94.8 & 94.1 \\
high & 94.8 & 94.4 & voice & 94.6 & 94.4 \\
back & 93.9 & 93.4 & continuant & 95.6 & 95.2 \\
low & 97.4 & 97.1 & nasal & 98.1 & 98.0 \\
anterior & 94.4 & 94.0 & strident & 97.8 & 97.6  \\
coronal & 94.3 & 93.9 & sil & 99.0 & 98.9 \\
\hline
\end{tabu}}
\end{table}

\begin{table} [t]
  \caption{\label{tab:eSPEaccuracies} {\it Classification accuracies (\%)
      of the eSPE analysers at frame level on cross-val. set
      of \textit{si\_tr\_s\_284} set.}}
\vspace{2mm}
\centerline{
\begin{tabu}{|l|c|c|[0.8pt]l|c|c|}
\hline
Phonological & \multicolumn{2}{c|[0.8pt]}{Accuracy (\%)} & Phonological &
\multicolumn{2}{c|}{Accuracy (\%)} \\
\cline{2-3} \cline{5-6}
features & train & cv & features & train & cv \\
\hline \hline
vowel & 94.7 & 94.3 & low & 97.5 & 97.2 \\
fricative & 97.3 & 97.0 & mid & 94.5 & 94.0 \\
nasal & 98.2 & 98.1 & retroflex & 98.6 & 98.4 \\
stop & 96.7 & 96.4 & velar & 98.9 & 98.8 \\
approximant & 97.2 & 96.9 & anterior & 94.8 & 94.3 \\
coronal & 94.8 & 94.4 & back & 94.2 & 93.8  \\
high & 94.6 & 94.2 & continuant & 95.8 & 98.4 \\
dental & 99.3 & 99.2 & round & 95.3 & 94.9 \\
glottal & 99.7 & 99.7 & tense & 91.4 & 90.7 \\
labial & 97.6 & 97.4 & voiced & 95.2 & 94.9 \\
\hline
\end{tabu}}
\end{table}






\subsection{Phonological synthesis}


\subsubsection{Training}
\label{sec:training}

The speech signals from the training and cross-validation sets of the
Tolstoy database, down-sampled to 16 kHz, framed by 25-ms windows with
10-ms frame shift, were used for extracting both DNN input and
output features. The input features, phonological posteriors
$\vec{z}_n$, were generated by the phonological analyser trained on the
WSJ database. The temporal context of 11 successive frames resulted in
input feature vectors of 132 ($12 \times 11 \times 1$), 165 ($15
\times 11 \times 1$) and 231 ($21 \times 11 \times 1$) dimensions, for
the GP, SPE and eSPE schemes, respectively. The output features, the
LPC speech parameters, were extracted by the Speech Signal Processing
(SSP) python
toolkit\footnote{\url{https://github.com/idiap/ssp}}. We used
static speech parametrization of 29th order along with its dynamic
features, altogether of 87th order. 

Cepstral mean normalisation of the input features was
applied before DNN training. The DNN was initialised using
$(K \times 11) \times 1024 \times 1024 \times 1024 \times 1024$ Deep
Belief Network pre-training by contrastive divergence with 1 sampling
step (CD1)~\citep{Hinton06}. The {\rv 4 hidden layers} DNN with a
linear output function was then trained using a mini-batch based
stochastic gradient descent algorithm with mean square error cost
function of the KALDI toolkit~\citep{PoveyASRU2011}. The DNN had 3.4
million parameters.

\subsubsection{Synthesis}

The test set of the Tolstoy database was used for the synthesis. There
are three options to generate a particular speech sound: (i) by $Z$
inferred from the speech, (ii) by $Z$ inferred from the text, or (iii)
by compositional speech modelling.
In the two first cases, we refer to this generation
as \emph{network-based}. The speech parameters generated by the
forward DNN pass are smoothed using dynamic features and pre-computed
(global) variances, and formant enhancement is performed 
to mitigate over-smoothing of the formant frequencies. 

In the latter case, speech sounds are generated using compositional
phonological speech models, introduced next in
Section~\ref{sec:soundsofprimes}. Briefly, we set a single
phonological feature as an active input and the rest are set to
zeros. This generates an artificial audio sound that characterises the
input phonological feature. Then, the speech sound is generated by the
composition (audio mixing) of the particular artificial audio
sounds. We refer to this process as \emph{compositional-based}. We
wanted a) to show that compositional phonological speech modelling is
possible, b) to test the suitability of speech sound synthesis without
the DNN, i.e., only by mixing  the artificial phonological sound
components, and c) to allow the reader to experiment with the sound
components that are embedded in this manuscript.



\section{Application examples for laboratory phonology}
\label{sec:examples}
In this section we show how to use the phonological vocoder, described
in Section~\ref{sec:experiments}, a) for compositional phonological
speech modelling (Section~\ref{sec:soundsofprimes}), b) for a comparison
of phonological systems (Section~\ref{sec:compphon}) and c) as a
parametric phonological TTS system (Section~\ref{sec:contsynth}).

\subsection{Compositional phonological speech models}
\label{sec:soundsofprimes}

\cite{Virtanen15} investigate the constructive compositionality of
speech signals, i.e., representing the speech signal as non-negative
linear combinations of atomic units (``atoms''), which themselves are
also non-negative to ensure that such a combination does not result in
subtraction or diminishment. The power of the sum of uncorrelated
atomic signals in any frequency band is the sum of the powers of the  
individual signals within that band. The central point is to define
the sound atoms that are used as the compositional models.

Following this line of research, we hypothesise that the acoustic
representation of the phonological features, produced by a
phonological vocoder, forms a set of speech signal atoms (the
phonological sound components) that define the phones. We
call these sound components \emph{phonological atoms}.
It is possible to generate the atoms for any phonological
system. 


\begin{figure}[ht]
\[
Z = \left(\begin{array}{*4{c}}
    \tikzmark{left}{1} & 0 & 0 & 0 \\
    0 & 1 & 0 & 0 \\
    0 & 0 & 1 & 0 \\
    \tikzmark{right}{0} & 0 & 0 & 1
  \end{array}\right)
  \Highlight[first]
\]

\centering
\begin{tikzpicture}[
    declare function={
      excitation(\t,\w) = sin(\t*\w);
      noise = rnd - 0.5;
      source(\t) = excitation(\t,20) + noise;
      filter(\t) = 1 - abs(sin(mod(\t, 50)));
      speech(\t) = 1 + source(\t)*filter(\t);
    }
]
  \draw[thick, ->] (1,5.5) -- (1,5);
  \draw[rounded corners, thick, fill=lightgray] (-0.2,3.6) rectangle (2.2,5);
  \node[align=center] at (1,4.25) {Synthesis\\DNN};

  \draw[thick, ->] (1,3.6) -- (1,3.1);

  \node[align=center] at (1.4,3.35) {\small{speech parameters}};
  \draw[rounded corners, thick] (-0.2,2) rectangle (2.2,3.1);
  \node[align=center] at (1,2.5) {LPC\\re-synthesis};

  \draw[thick, ->] (1,2) -- (1,1.5);

\draw[blue, thick, x=0.0085cm, y=0.4cm] (0,1) -- plot [domain=0:360,
samples=144, smooth] (\x,{speech(\x)});

\end{tikzpicture}
  \caption{{ \it Generation of a sound of the first phonological atom. The
      artificial phonological representation, the identity matrix, is
      for illustration of size 4. {\rv The phonological synthesis DNN
      generates the speech parameter vectors: the 24th order line
      spectral pairs, the harmonic to noise ratio, and the angle and magnitude
      of a glottal pole. The speech parameters are then
    re-synthesized by LPC re-synthesis.}}}
\label{fig:sop}
\end{figure}
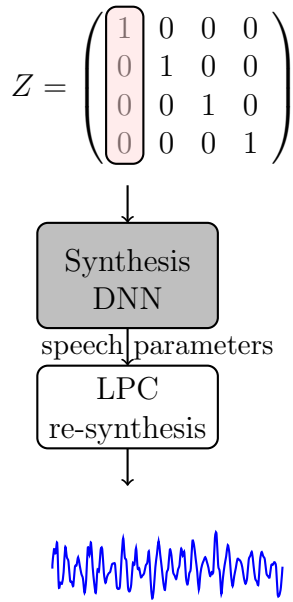

Given our hypothesis above (that these phonological atoms form a set
of acoustic templates that might be taken to define speech acoustic
space), we defined artificial representations
$Z=\{\vec{z}_1,\ldots,\vec{z}_k,\ldots,\vec{z}_K\}$ as the identity 
matrix $I_K$ of size $K \times K$. Each column represents a
phonological atom, and its speech samples are generated as described
in Section~\ref{sec:resynthesis}, and illustrated in
Figure~\ref{fig:sop}.

Thus, we can listen {\rv to} the sounds of individual phonological
features. As an example, Table~\ref{tab:SOPaudio} in \ref{sec:samples}
demonstrates recordings of phonological features of the GP system. We
generated these \emph{phonological sounds} also for the SPE and eSPE
phonological systems and used all of them in the following
experiments.

Finally, the composition of the speech sounds, the phones, driven by
the canonical phone representation, is done as follows:
\begin{equation}
  \label{eq:fusion}
  \vec{y}_n=\frac{1}{S}\sum_{s=1}^S w_n^s z_n^s,
\end{equation}
where $\vec{y}_n$ is a composition of $S$ (a particular subset of $K$
phonological features) phonological atoms, and $w_n^s$ are weights of
the composition. The weights are a-priori mixing coefficients not
investigated closer in this work; we used them as constants with the
value of 1.


{\rv The presented compositional speech sound generation is
  context-independent and generates the middle parts of the phones.}

\subsection{Comparison of the phonological systems}
\label{sec:compphon}

We start with context-independent vocoding in
Section~\ref{sec:civ}, i.e., the vocoding of isolated speech sounds,
and continue with context-dependent vocoding in Section~\ref{sec:cdv}.

\subsubsection{Context-independent vocoding}
\label{sec:civ}

The aim of this subsection is to objectively evaluate the phonological
systems in respect to context-independent vocoding, i.e., the
ability of phonological systems to produce isolated speech sounds.
In order to achieve this, instead of inferring the posteriors from the
speech, we generated the canonical phonological posteriors, i.e., we
used only the features that represent the specific isolated phones
(rows of Tables \ref{tab:GPatts}, \ref{tab:SPEatts}, and
\ref{tab:eSPEatts}). Original speech, manually phonetically labelled
76 utterances from the audiobook test set, was used as the reference
for the comparison.
Having the phoneme boundaries, we extracted 25ms windows
from the central (stationary) part of the phones, to obtain human
spoken acoustic references.

\begin{figure}[ht]
  \centering
  \includegraphics[width=.73\linewidth]{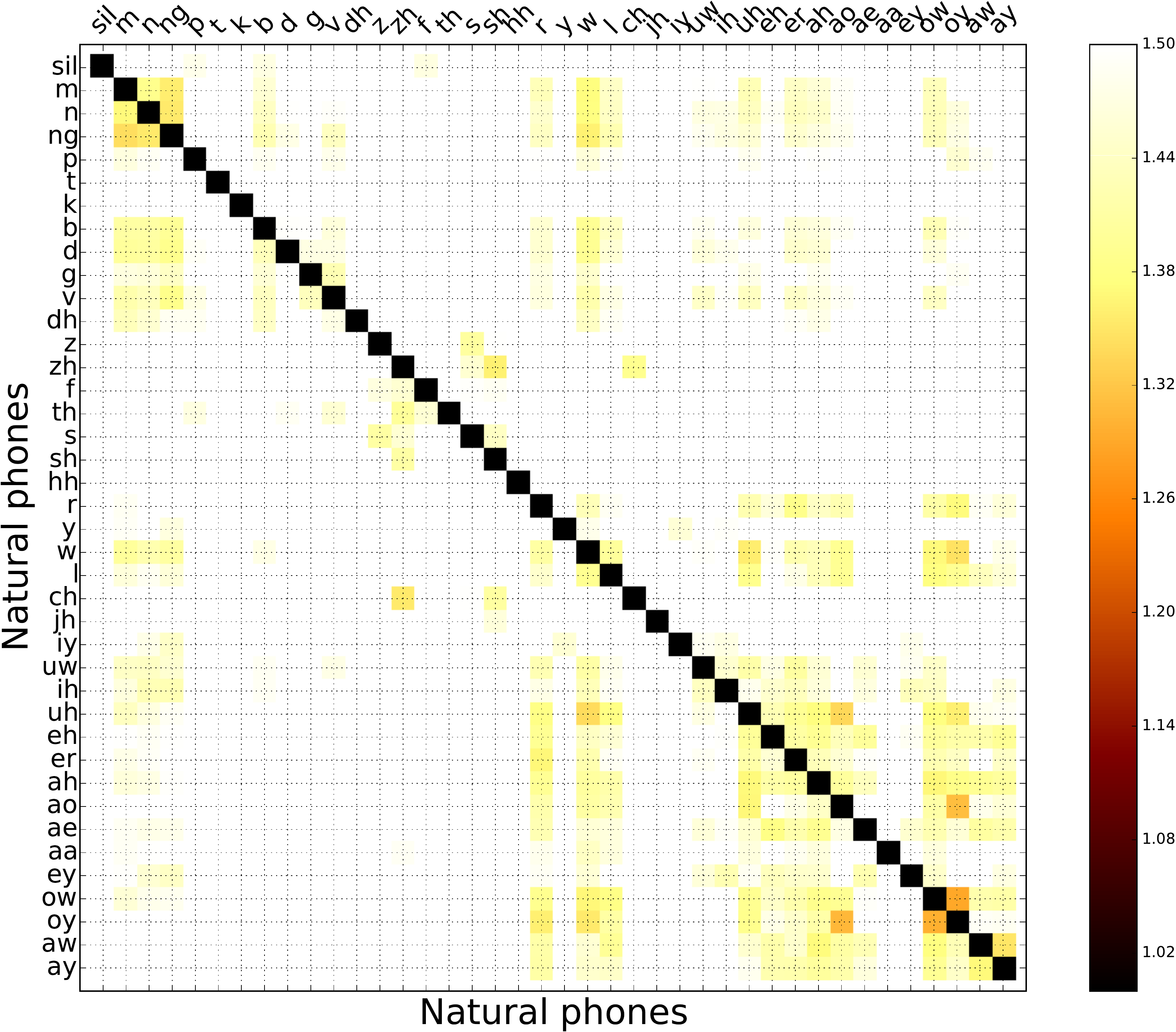}
  \caption{{\it Scaled distance matrix $D_{natural}$ of the spoken
      acoustic references.}}
\label{fig:human-mcd}
\end{figure}

Finally, we used the Mel Cepstral Distortion (MCD)~\citep{Kubichek93}
to calculate perceptual acoustic distances between the $P$ vocoded
phones and the spoken references, a distance matrix $D_{vocoded}$.
The MCD values are in dB and higher values represent more confused
phones. In order to visually compare confusions introduced by the
different phonological representations, we normalised $D_{vocoded}$ by
distances between the spoken references only, a distance matrix
$D_{natural}$, resulting into the distance matrix $D_{norm}$. The
definitions of the distance matrices are as follow: 


\begin{align}
\label{eq:scalingC}
  D_{natural} &= [D_1,\ldots,D_p,\ldots,D_P] \\
  D_p &= \vec{1} + \left(\vec{1} - \left(\max\left(N_p\right)\vec{1} - N_p\right)\oslash\left(\max\left(N_p\right)\vec{1} - \min\left(N_p\right)\vec{1} \right) \right)\\
  D_{norm} &= D_{vocoded} \odot D_{natural}
\end{align}
where $\vec{1}$ is the all ones vector of dimension $P$, $N_p$ is a
vector of MCDs values between a phoneme $p$ and all other phonemes,
and $D_p$ are the scaled values between $1$ and $2$, by $1$ referring
to the minimal distance (no confusion). 
Operator $\odot$ stands for element-wise matrix multiplication (the
Hadamard product), $\oslash$ stands for element-wise matrix division.
For illustration, Figure~\ref{fig:human-mcd} shows the distance matrix
$D_{natural}$. {\rv To better visualise the contrast of the distances
around 1, the maximum value of the colour map is set to 1.5 (i.e., the
distances above 1.5 are also white).}

\begin{figure}
\centering
\begin{subfigure}{.5\textwidth}
  \centering
  \includegraphics[width=1.\linewidth]{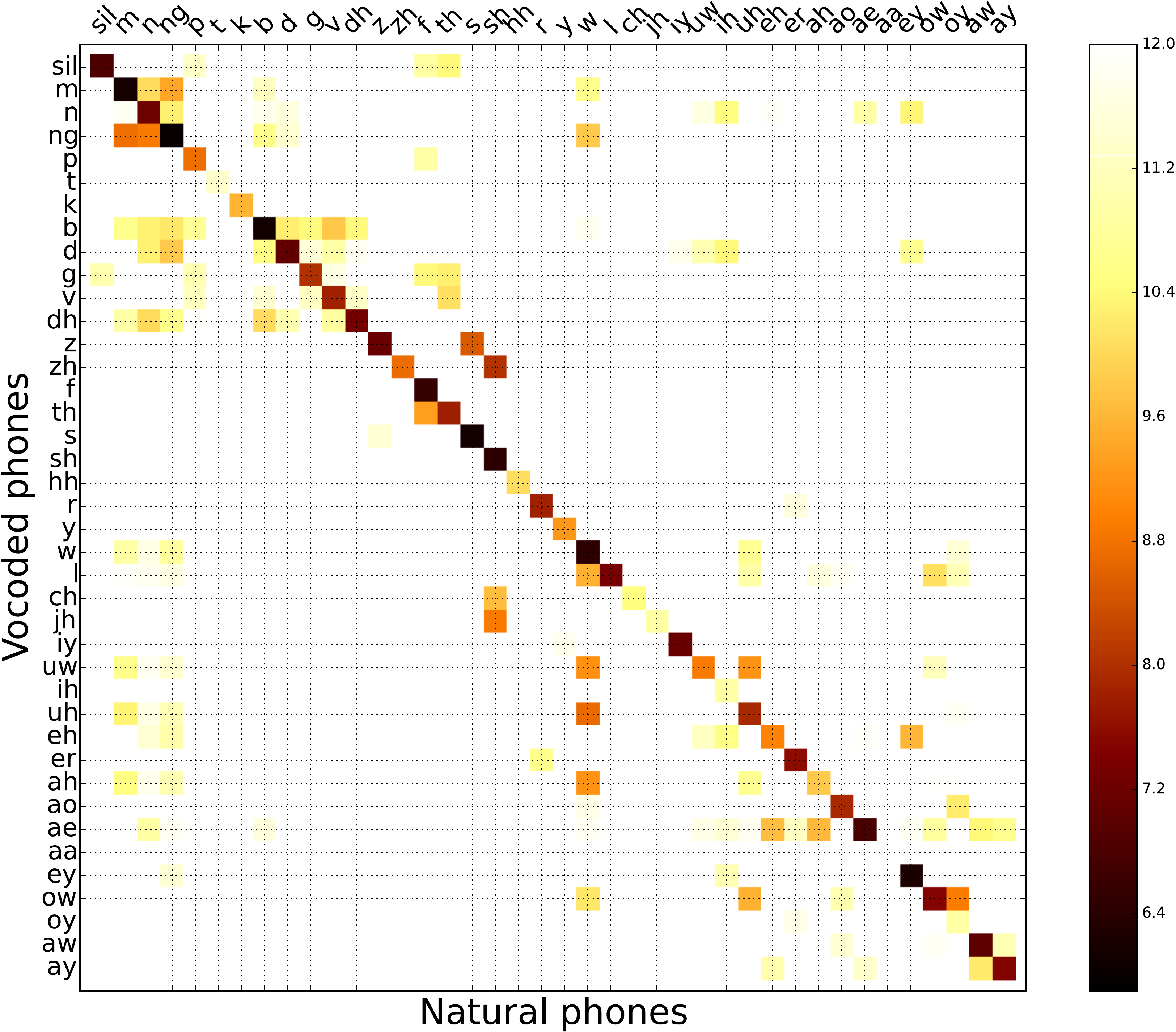}
  \caption{Network-based}
  \label{fig:GPsoundsPruned}
\end{subfigure}%
\begin{subfigure}{.5\textwidth}
  \centering
  \includegraphics[width=1.\linewidth]{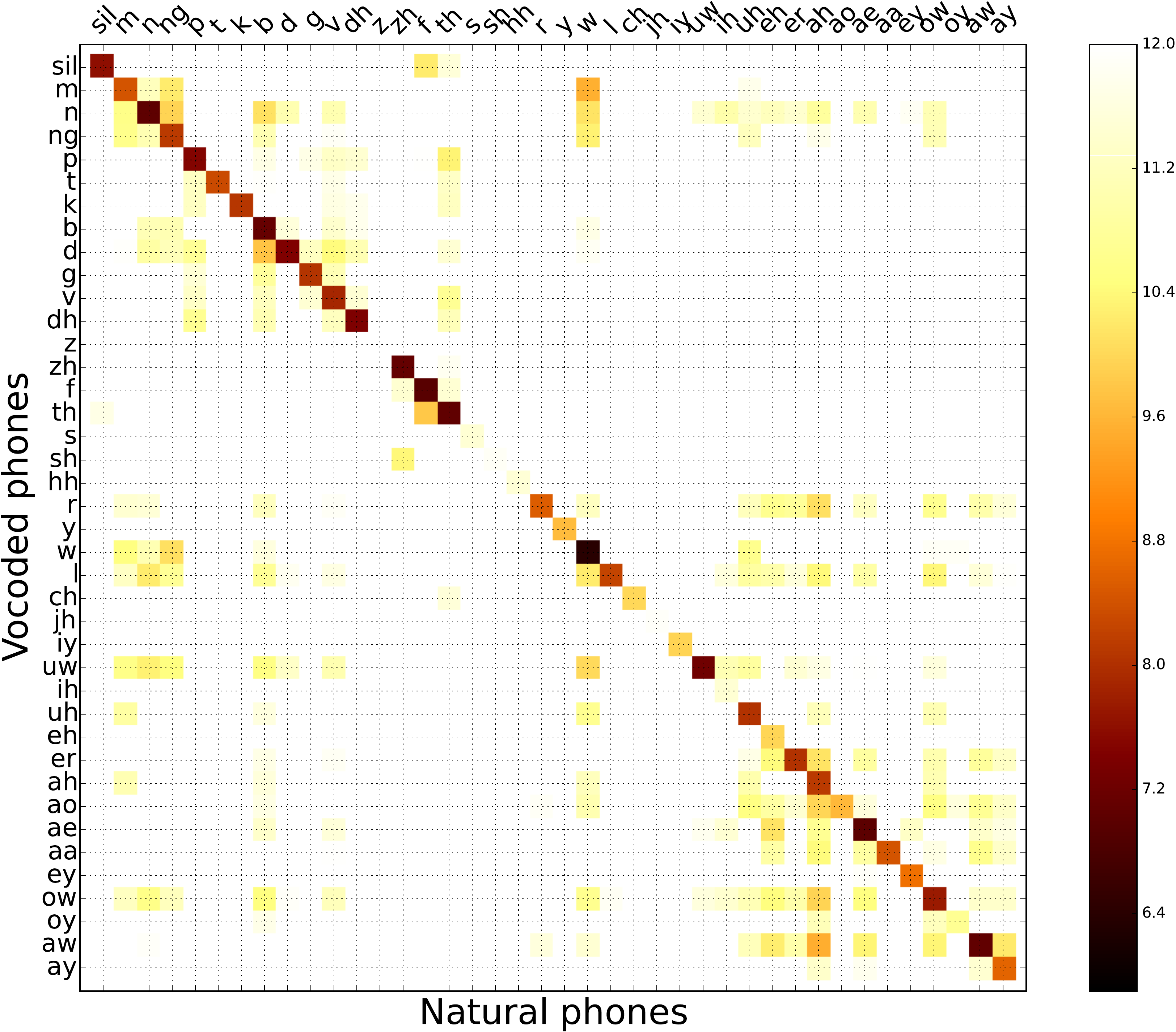}
  \caption{Compositional-based}
  \label{fig:GPsoundsCombined}
\end{subfigure}
\caption{{\it $D_{norm}$ of  GP vocoded and spoken phones.}}
\label{fig:GPSoundTests}
\end{figure}

\begin{figure}
\centering
\begin{subfigure}{.5\textwidth}
  \centering
  \includegraphics[width=1.\linewidth]{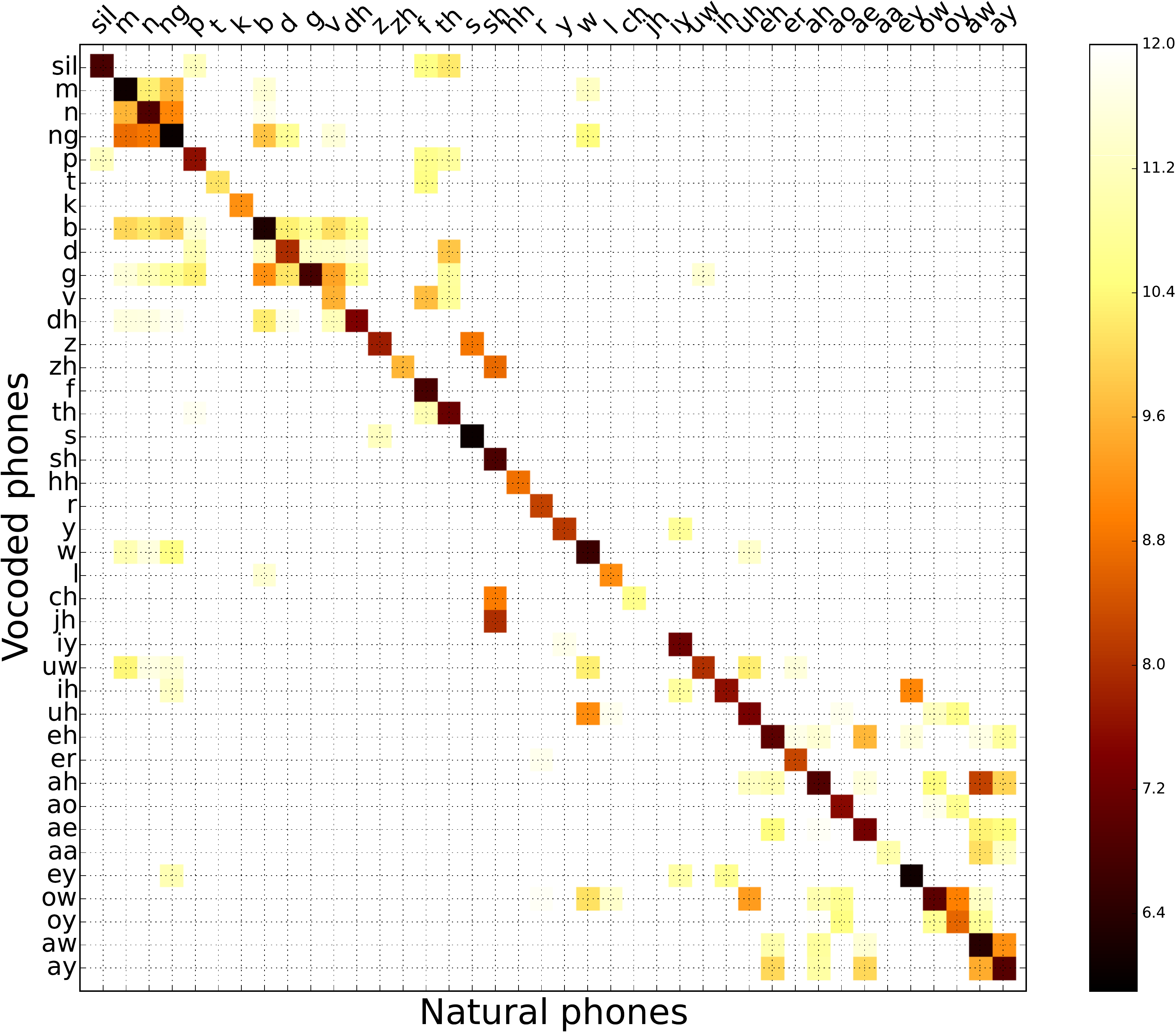}
  \caption{Network-based}
  \label{fig:SPEsoundsPruned}
\end{subfigure}%
\begin{subfigure}{.5\textwidth}
  \centering
  \includegraphics[width=1.\linewidth]{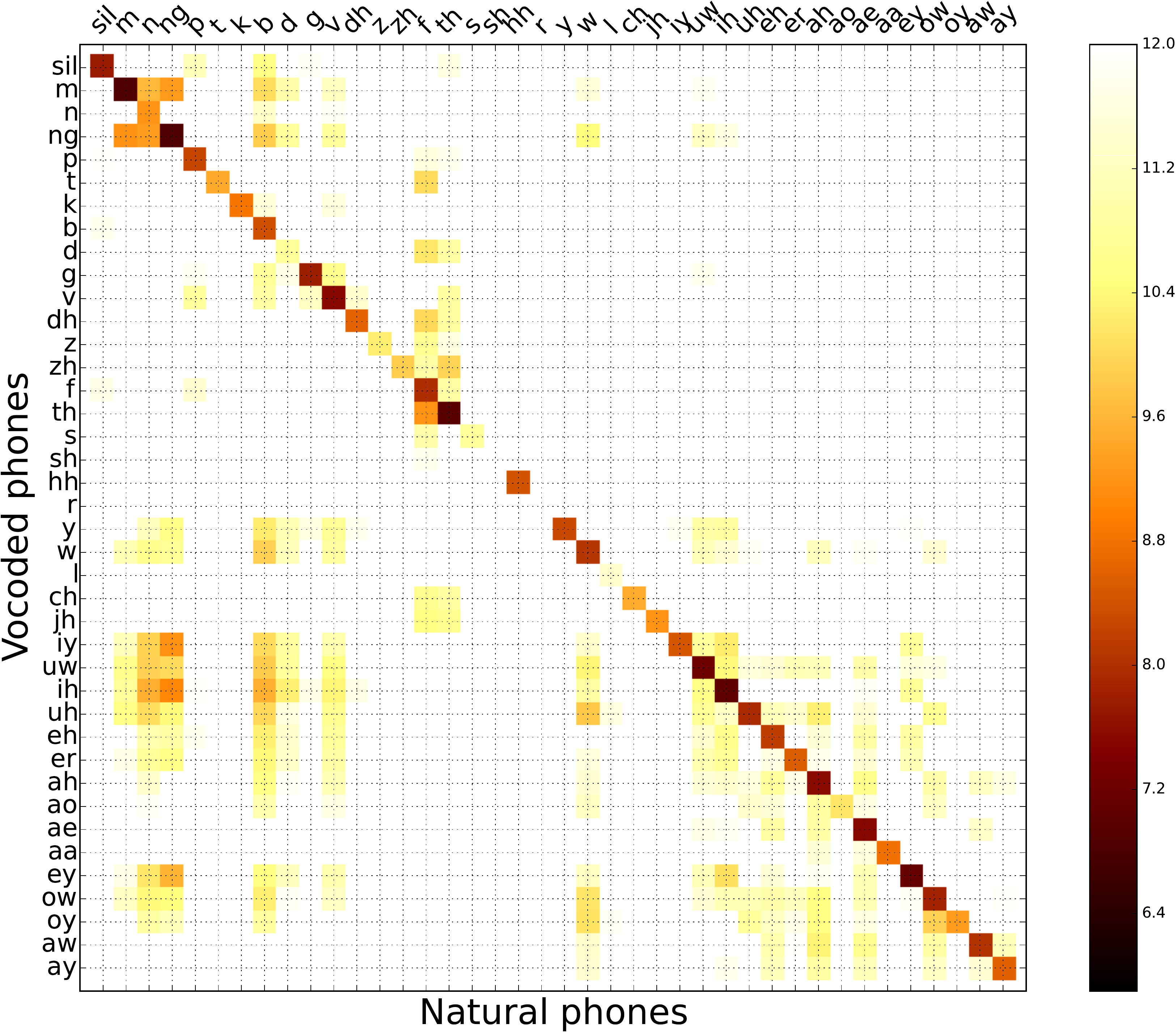}
  \caption{Compositional-based}
  \label{fig:SPEsoundsCombined}
\end{subfigure}
\caption{{\it $D_{norm}$ of SPE vocoded and spoken phones.}}
\label{fig:SPESoundTests}
\end{figure}

\begin{figure}
\centering
\begin{subfigure}{.5\textwidth}
  \centering
  \includegraphics[width=1.\linewidth]{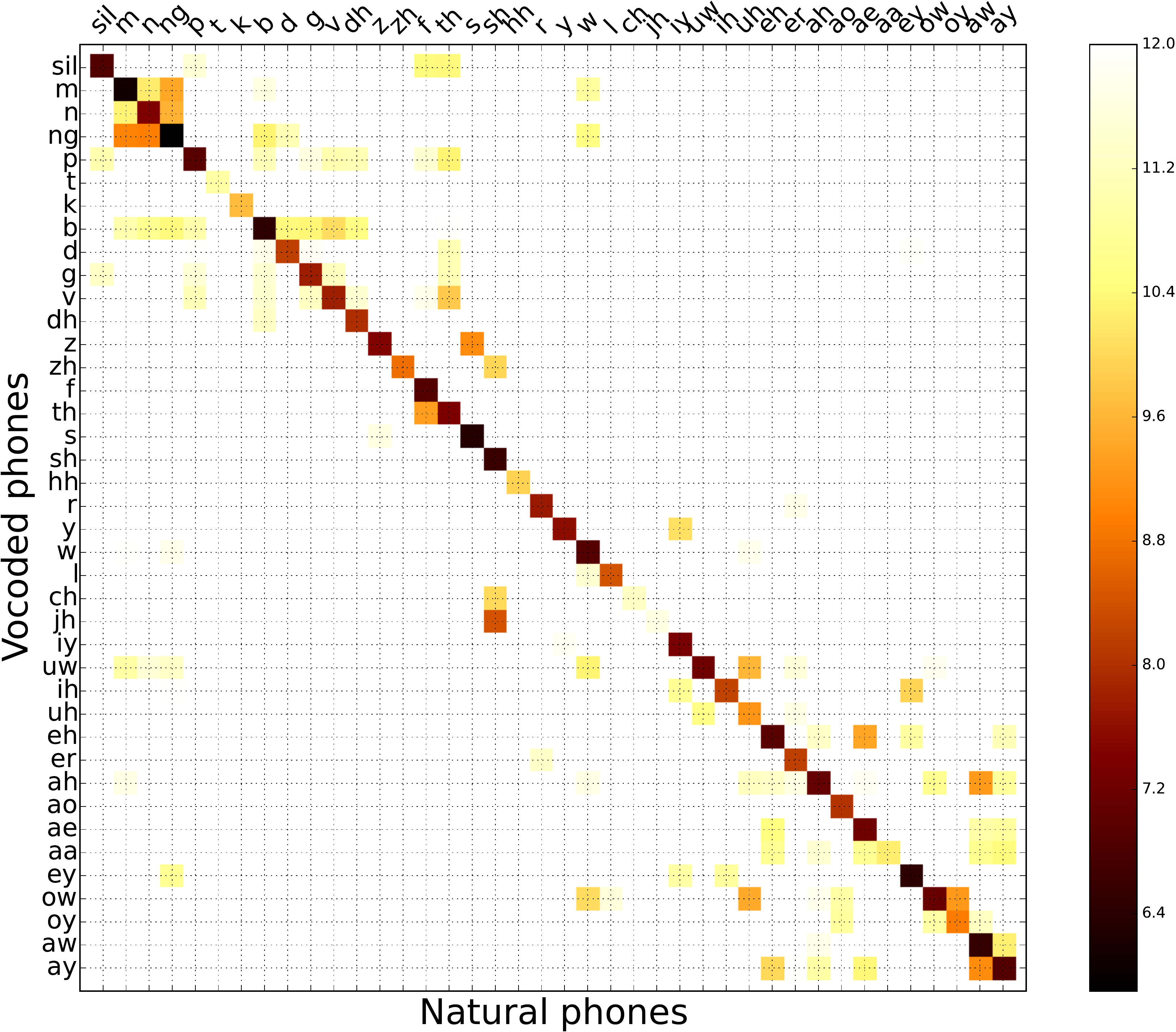}
  \caption{Network-based}
  \label{fig:eSPEsoundsPruned}
\end{subfigure}%
\begin{subfigure}{.5\textwidth}
  \centering
  \includegraphics[width=1.\linewidth]{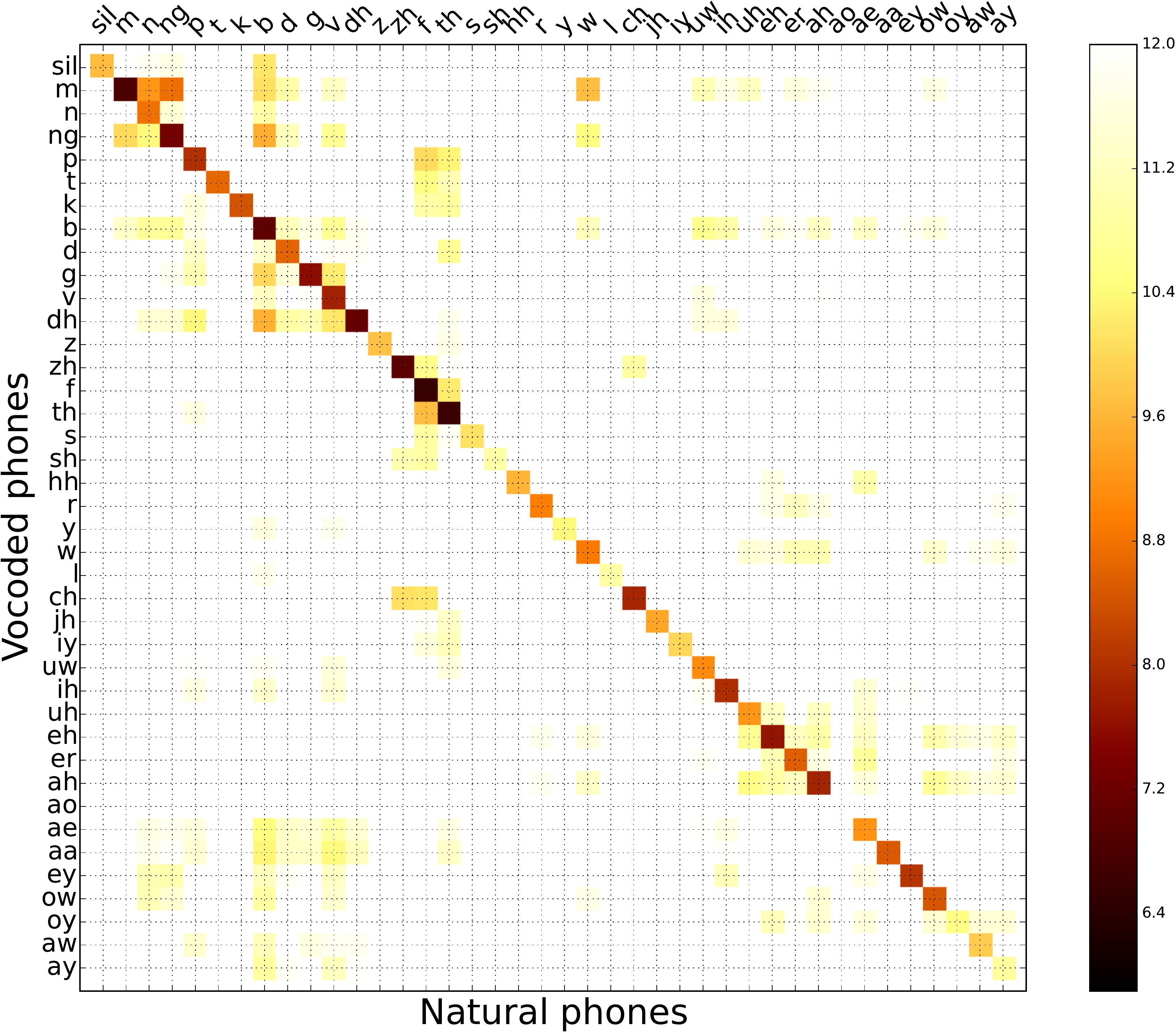}
  \caption{Compositional-based}
  \label{fig:eSPEsoundsCombined}
\end{subfigure}
\caption{{\it $D_{norm}$ of eSPE vocoded and spoken phones.}}
\label{fig:eSPESoundTests}
\end{figure}

Figures~\ref{fig:GPSoundTests}, \ref{fig:SPESoundTests}
and~\ref{fig:eSPESoundTests} show normalised distance matrices
$D_{norm}$ of vocoded context-independent phones and the spoken
acoustic references of the same speaker, for the GP, SPE and eSPE
phonological systems, respectively. The diagonal elements of the
distance matrices represent an acoustical distance (dissimilarity) of
vocoded and spoken phones. If the phonological features represent
speech well, the matrices show only strong diagonals. Missing
diagonal values imply greater distance between spoken and vocoded
phones, and might be caused by erroneous assignment of phonological
features to phonemes (the maps given in \ref{sec:spefeatures}) during
the training of the phonological analysis.


The figures show two aspects of the evaluation. In the first, the
distance matrices of (a) the network-based and (b) the
compositional-based phonological synthesis are shown.
The network- and compositional-based synthesis differ in the way
features are combined. In the first case, the DNN inputs are combined
so as to generate specific phones. In the second case, phonological
atoms of the specific phone are combined in order to generate
compositional phones. We can therefore  consider this as two different
evaluation metrics, hypothesising that both contribute partially to a
final evaluation. In both cases, the ideal performance is to have dark
diagonals, i.e., the lowest acoustic distance between the spoken
reference and vocoded phones. We consider the deviations from the
ideal performance as the errors and this allows us to search for
patterns in these errors. Table~\ref{tab:diagonals} shows the averages
of the diagonal values of the distance matrices.

\begin{table} [htb]
  \caption{\label{tab:diagonals} {\it The average MCD values given in
      [dB] of the diagonals of the distance matrices shown in
      Figures~~\ref{fig:GPSoundTests}, \ref{fig:SPESoundTests}
      and~\ref{fig:eSPESoundTests}.}}
\vspace{2mm}
\centerline{
\begin{tabular}{|r|c|c|c|}
\hline
System & Network & Compositional \\
\hline
GP & 8.12 & 8.67 \\
\hline
SPE & 7.79 & 8.77 \\
\hline
eSPE & 7.87 & 8.79 \\
\hline
\end{tabular}}
\end{table}

%

For all three phonological systems, the results of
compositional-based context-independent vocoding show greater errors
(i.e., higher values in Table~\ref{tab:diagonals}) than the ones of
network-based vocoding. This might be caused by the fact that
the composition of the phonological atoms is a linear operation (as shown
in Section~\ref{sec:soundsofprimes}), and it is an approximation to a
non-linear function that is modelled by network-based phonological
synthesis. Additionally, the types of reported {\rv errors} in the
network-based vocoding that are missing in the compositional-based one
make sense phonetically. For example, in the left panel of
Figure~\ref{fig:GPsoundsPruned}, phonetically very similar [\ph{Z}],
[\ph{S}], [\ph{dZ}], [\ph{tS}] with voicing and closure being highly
context dependent {\rv have also smaller acoustic
  distances}. Nevertheless, compositional-based vocoding tends to
produce {\rv smaller errors} in all three frameworks. For example,
while [\ph{D}] in the left panel of Figure~\ref{fig:GPsoundsPruned}
{\rv has smaller acoustic distances with nasals, voiced plosives and
[\ph{v}], the right panel shows only smaller distances only with labials.}

In the second aspect of the evaluation, the three phonological systems
(GP, SPE and eSPE) are compared among
themselves using network-based synthesis. In
Figures~\ref{fig:GPsoundsPruned}, \ref{fig:SPEsoundsPruned}, and
\ref{fig:eSPEsoundsPruned}, we see different error patterns. In all
three phonological systems the biggest {\rv errors} are shown with the
nasals [\ph{m n N}]. GP in addition produces {\rv errors in} the
consonants and vowels, such as for nasals. SPE seems to represent
speech better, namely for vowel [\ph{A}] and glide [\ph{w}], and
suppresses most of the vowel-consonant {\rv errors}. On the other hand,
it fails with proper [\ph{dZ}] vocoding. We speculate that phone
frequency in the evaluation data may be one of the causes of these
errors; for example, the phone [\ph{dZ}] was the least frequent one in
our data.
Finally, according to our data, eSPE further improves on the SPE
speech representation. It generates fewer {\rv errors} in the vowel
space, and also in the consonant space, for example in the voiced
stops class [\ph{b d g}].


\subsubsection{Context-dependent vocoding}
\label{sec:cdv}

The previous experiment evaluated the vocoding of the isolated
sounds, using canonical posteriors $\vec{z}_n$. We continued with the
evaluation of continuous speech vocoding using $\vec{z}_n$ inferred
from the reference speech signals. Network-based phonological
synthesis was used for the following experiments.

In this evaluation, we were interested if the segmental errors found
in context-independent vocoding impact the context-dependent
vocoding. { \rv We employed a subjective evaluation listening
  test~\citep{Loizou2011}, suitable for comparing two different
  systems. In this pair-wise comparison test, listeners were presented
  with pairs of samples produced by two systems and for each pair they
  indicated their preference. The listeners also were presented with
  the choice of ``no preference'', when they couldn't perceive any
  difference between the two samples. The material for the test
  consisted of 16 pairs of sentences such that one member of the pair
  was generated using the GP-based vocoder and the other member was
  generated using the eSPE-based vocoded speech. }
Random utterances from the test set of the Tolstoy database were used
to generate the vocoded speech.
We chose these two systems because they displayed the greatest
differences in context-independent results. The subjects for {\rv the
  listening test} were 37 listeners, roughly equally pooled from
experts in speech processing on the one hand, and completely naive
subjects on the other hand. The subjects were presented with pairs of
sentences in a random order with no indication of which system they
represented. They were asked to listen to these pairs of sentences (as
many times as they wanted), and choose between them in terms of their
overall quality. {\rv Additionally, the option ``no preference'', was
  available  if they had no preference for either of them.} To
decrease the demands on the listeners, we divided the material into
two different sets, each consisting of 8 paired sentences randomly
selected from the test set. The first set was presented to 19
listeners, and the second set to 18 different listeners.

\begin{figure}[h]
\centering
\includegraphics[width=.8\linewidth]{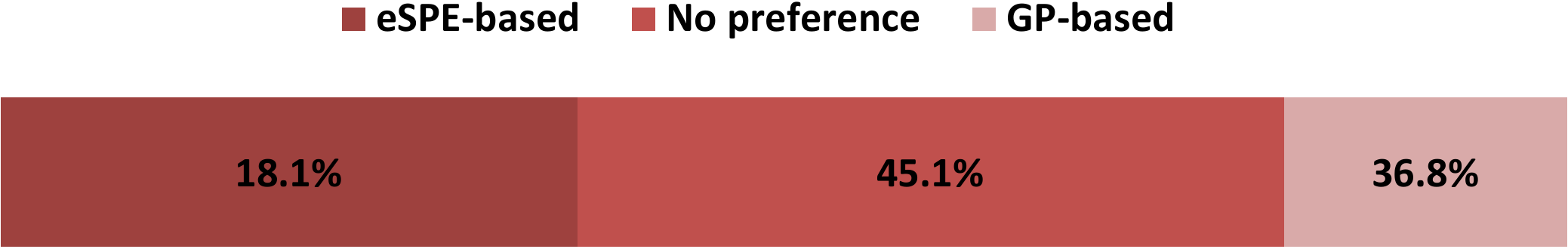}
\caption{{\rv Subjective preference score (\%)} between the GP-based
and the eSPE-based vocoded speech}
\label{fig:SubEvalGP_eSPE}
\end{figure}

In Figure~\ref{fig:SubEvalGP_eSPE} the results of {\rv the subjective
  listening test} for the GP and eSPE phonological systems, are
shown. As can be seen, the eSPE-based phonological vocoder outperforms
the GP one by 36.8\% compared with 18.1\% preference score. Even though
there is a preference of the listeners towards the eSPE-based system
(double preference percentage), it is clearly shown that with a very
high percentage, 45\%, the two systems are perceived by the listeners
as having the same overall quality {\rv (\emph{no preference})}. 
It should be pointed out that a
$t$-test confirmed that this difference between the GP-based and
eSPE-based phonological vocoders is statistically significant ($p <
0.01$). 

We hypothesize that the preference for eSPE is linked to
greater perceptual clarity of individual phones. Subsequent auditory
and visual analyses of the sample sentences and their generated
acoustic signals suggest that eSPE sentences displayed longer closures
for plosives, stronger plosive releases, and also slightly greater
disjunctures at some word boundaries.
These features correspond to a decreased overlap of sounds,
i.e. decreased coarticulation, commonly present in hyper-articulated
or clearly enunciated speech.



\subsection{Experimental parametric phonological TTS}
\label{sec:ptts}

In this section we show how compositional phonological speech models
could be combined to generate arbitrary speech sounds, and how to
synthesise continuous speech from the canonical phonological
representation.

\subsubsection{Generation of sounds from unseen language}

In this experiment, we arbitrarily selected the GP system to demonstrate
the phonological composition of new speech sounds. \cite{Harris94}
claims that fusing and splitting of primes accounts for phonological
description of the sound. We selected the phonological rule number 29
[I, U, E] $\rightarrow$ \ph{y}, and  [A, I, U, E] $\rightarrow$
\ph{\oe}, of \cite{Harris94} and tried to synthesise non-English
sounds by the composition of involved phonological atoms. According to
Section~\ref{sec:soundsofprimes}, we claim that new sounds can be
generated by time-domain mixing of the corresponding atoms
$\vec{a}^s$. Table~\ref{tab:SOPcombaudio} demonstrates the synthesis of
standard German sounds [\ph{y}] and [\ph{\oe}] from English
phonological atoms $\vec{a}^s$, generated as in
Eq.~\ref{eq:fusion}. For $w_n^s=1$, it can be done easily with
available free tools, e.g.:
\begin{itemize}
  \item [[\ph{y}]] :  \texttt{sox -m I.wav U.wav E.wav y.wav}
  \item [[\ph{\oe}]] : \texttt{sox -m A.wav I.wav U.wav E.wav oe.wav}
\end{itemize}

\begin{figure}[h]
\centering
\includegraphics[width=.8\linewidth]{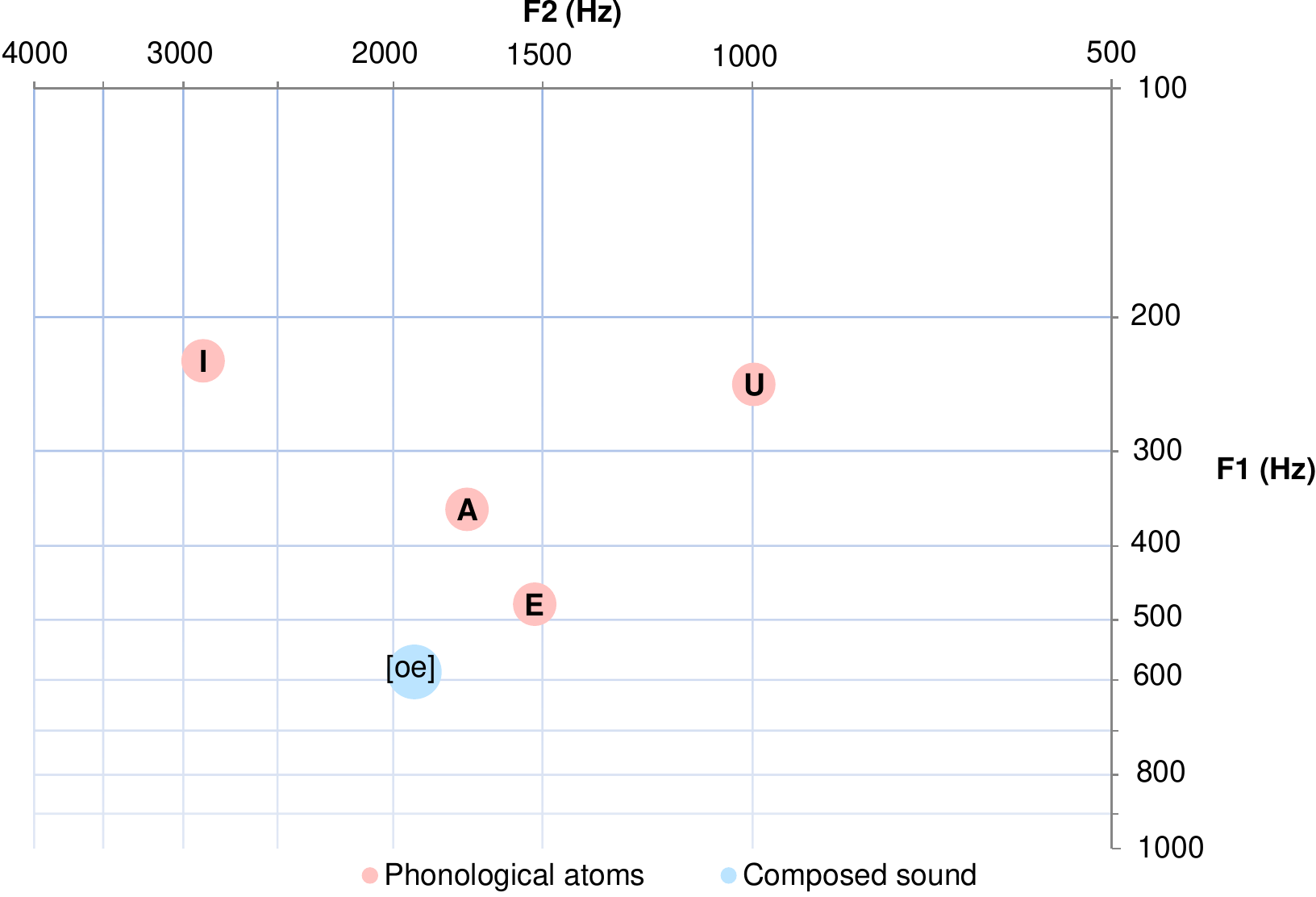}
\caption{First two formants of involved phonological atoms and their
  composition -- a phone [\ph{\oe}].}
\label{fig:formants}
\end{figure}

We performed a formant analysis of all phonological atoms, and concluded
that they contain the same number of formants as human speech sounds
(i.e., 4 in the 5 kHz bandwidth). In addition, the combined sounds
also contain the proper number of formants. The first two formants play a
major perceptual role in distinguishing different English
vowels~\citep{Ladefoged14}, and Figure~\ref{fig:formants} shows
F1 and F2 of [\ph{\oe}] phone from
Table~\ref{tab:SOPcombaudio}. The mean value of the first formant of
the natural [\ph{\oe}] takes values between 509 Hz~\citep{Gendrot05}
and 550 Hz~\citep{patzoldSimpson97}. The mean values of the second
formant of the natural [\ph{\oe}] takes values between 1650 Hz and
1767 Hz. The average F1 of synthesized [\ph{\oe}] is 559 Hz and F2 is
1917 Hz. Those numbers are slightly higher than average population
numbers, however, they may be related to our female speaker used in
the training of the phonological synthesis.

The results of this experiment support the hypothesis mentioned in
Section  \ref{sec:soundsofprimes}, that phonological atoms may define
the phones. In addition,  Section \ref{sec:civ} demonstrated that
compositional-based phonological synthesis works well also for the SPE
and eSPE phonological systems.

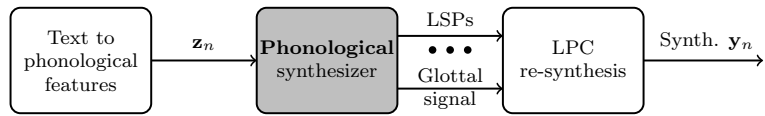
\begin{figure}[!th]
\centering
\resizebox{4in}{!}{%
\begin{tikzpicture}[font=\scriptsize]
  \draw[rounded corners, thick] (0,0) rectangle (2,1.5);
  \node[align=center] at (1,0.75) {Text to\\phonological\\features};

  \draw[thick, ->] (2,0.75) -- (3.5,0.75); \node[above] at (2.75,0.75)
       {$\vec{z}_n$};
  \draw[rounded corners, thick, fill=lightgray] (3.5,0) rectangle (5.5,1.5);
  \node[align=center] at (4.5,0.75) {\textbf{Phonological}\\synthesizer};

  \draw[thick, ->] (5.5,1.1) -- (7,1.1); \node[above] at (6.25,1.1)
       {LSPs};

  \draw[thick, fill=black] (6,0.9) circle[radius=0.04];
  \draw[thick, fill=black] (6.25,0.9) circle[radius=0.04];
  \draw[thick, fill=black] (6.5,0.9) circle[radius=0.04];

  \draw[thick, ->] (5.5,0.35) -- (7,0.35); \node[align=center] at (6.25,0.35)
       {Glottal\\signal};

  \draw[rounded corners, thick] (7,0) rectangle (9,1.5);
  \node[align=center] at (8,0.75) {LPC\\re-synthesis};
  \draw[thick, ->] (9,0.75) -- (10.7,0.75); \node[above] at (9.9,0.75)
       {Synth. $\vec{y}_n$};
\end{tikzpicture}
}
\caption{Phonological TTS synthesis. Speech parameters, the speech
  line spectral pairs LSPs and source parameters, are generated by the
  DNN. Speech samples are generated by subsequent LPC re-synthesis.}
\label{fig:phonovocTTS}
\end{figure}

\subsubsection{Continuous speech synthesis}
\label{sec:contsynth}

The composition of Eq.~\ref{eq:fusion} represents a static mixing of
$S$ phonological atoms, i.e., it cannot be applied to model
co-articulation. To include co-articulation into the synthesis, the
phonological synthesizer has to be used. As it was trained with the
temporal context of 11 successive frames, around 50 ms before and 50
ms after the current processing frame, it learnt how speech parameters
change with trajectories of the phonological posteriors.

Experimental parametric phonological TTS can be designed by a
simplistic text processing front end: input text transformed into the
phonemes using a lexicon, and the phonemes transformed to phonological
features using maps given
in~\ref{sec:spefeatures}. Figure~\ref{fig:phonovocTTS} shows the TTS
process with the phonological synthesis. The binary phonological
representation to be synthesized is obtained again from the canonical
phone representation.

To demonstrate the potential of our parametric phonological TTS
system, we randomly selected three utterances from a \texttt{slt}
subset of the CMU-ARCTIC speech database~\citep{Kominek2004}, and used
their text labels to generate continuous speech. Specifically, we used
the phoneme symbols along with their durations from the forced-aligned
full-context labels provided with the database, and mapped it to the
phonological representation. Then we synthesised the sentences using
the already trained phonological synthesizers as described in
Section~\ref{sec:training}.

\begin{figure}
\centering
\begin{subfigure}{1.0\textwidth}
  \centering
  \includegraphics[width=.97\linewidth]{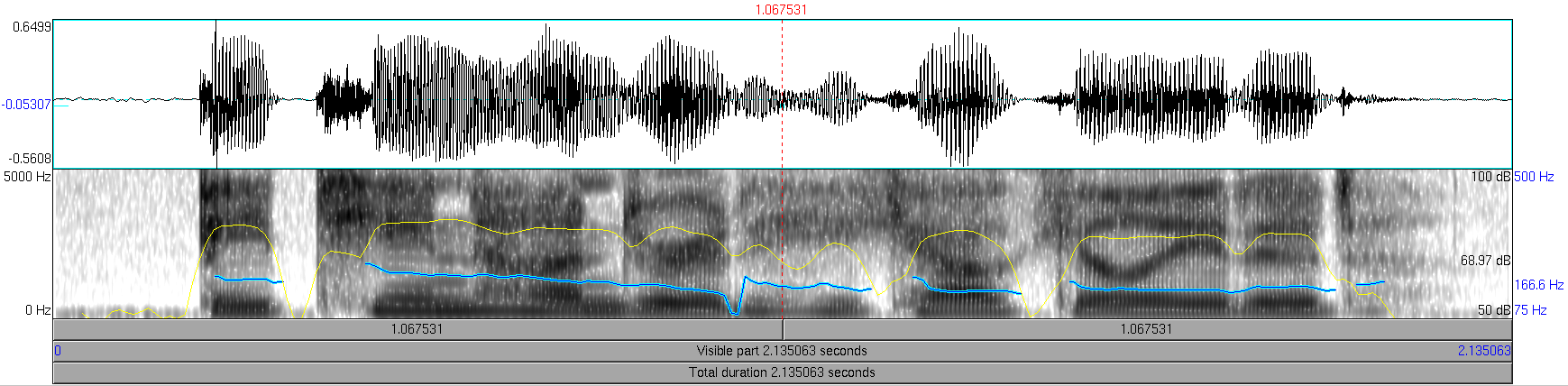}
  \caption{The natural speech of the arctic\_a0453 example.}
  \label{fig:a543-orig}
\end{subfigure}%
\\
\begin{subfigure}{1.0\textwidth}
  \centering
  \includegraphics[width=.97\linewidth]{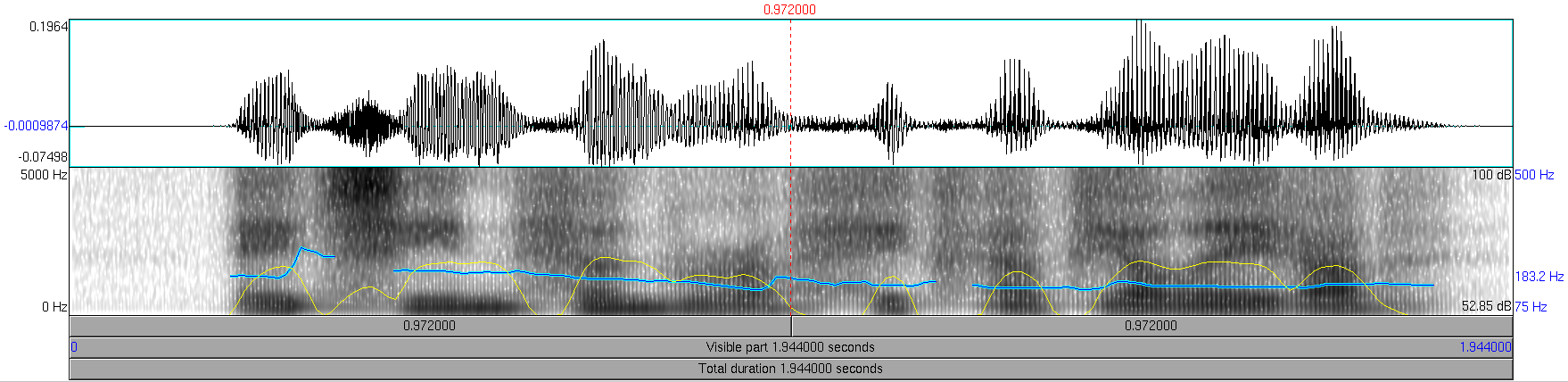}
  \caption{The example generated with the GP features.}
  \label{fig:a543-GP}
\end{subfigure}%
\\
\begin{subfigure}{1.0\textwidth}
  \centering
  \includegraphics[width=.97\linewidth]{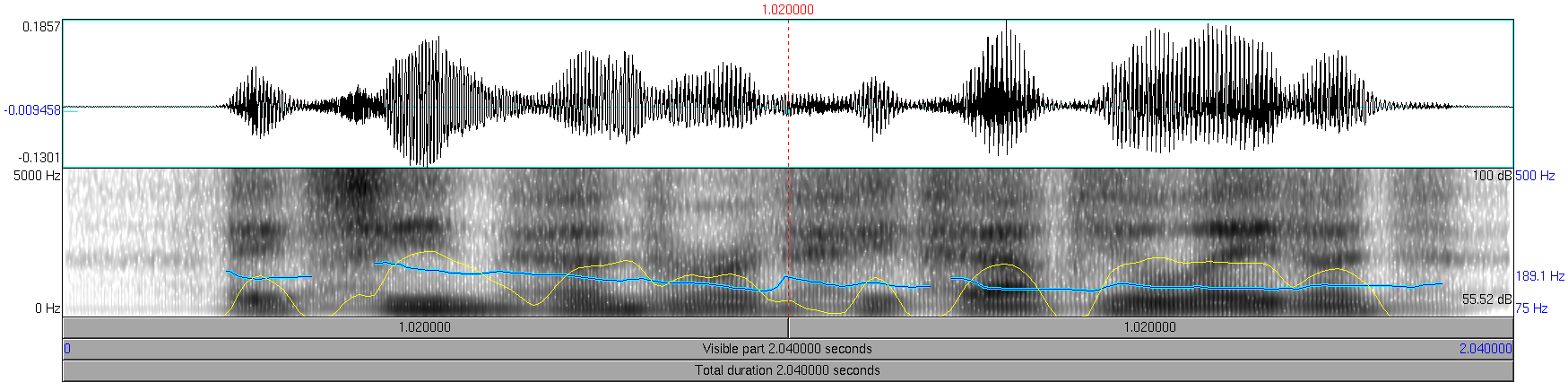}
  \caption{The example generated with the eSPE features}
  \label{fig:a543-eSPE}
\end{subfigure}
\caption{{\it Visualisation of vocoded arctic\_a0453 examples:
    ``Eighteen hundred, he calculated.''. The GP vocoding seems to
    better synthesise the higher frequencies such as for fricatives,
    whereas eSPE vocoding seems to synthesise stronger formant
    frequencies. The recordings are available in \ref{sec:samples}.}}
\label{fig:a543-examples}
\end{figure}

Table~\ref{tab:SPEaudio} lists recordings that demonstrates speech
synthesis from the phonological speech representation. The example
\texttt{a0453} illustrates how the phonological
vocoder learns the context. Figure~\ref{fig:a543-examples} visualises
the generated GP and eSPE examples. The phoneme sequence of the first
word is \ph{[eI t i n]}, while the synthesised sequences using both
phonological systems are rather \ph{[eI tS i n]}. The substitution of
\ph{[t]} by \ph{[tS]} illustrates the assimilation of the place of
articulation in the synthesised phoneme \ph{[t]}. If \ph{[t]} starts a
stressed syllable and is followed by \ph{[i]} this alveolar stop is
aspirated and commonly more palatal due to coarticulation with the
following vowel. The acoustic result of a release burst when the
tongue is in the alveo-palatal region is similar to the frication
phase of alveo-palatal affricate \ph{[tS]}. 
We also observe that the stop closure in the synthesized speech is
much shorter and not realized as complete silence, which is consistent
with the substitution of \ph{[t]} by \ph{[tS]} but may also be related
to other general settings for the synthesis.
We conclude that the phonological synthesizer learns some contextual
information because of using the temporal window of 11 successive
frames -- around 100 ms of speech, that may correspond to the formant
transitions and differences in voice onset times. This is probably
enough to learn certain aspects of co-articulation well.


\paragraph{Subjective Intelligibility Test}

We compared the phonological TTS with a conventional hidden Markov
model (HMM) parametric speech synthesizer, trained on the same
training set of the audiobook which was used for training the
phonological vocoder. For building the HMM models, the HTS V.2.1
toolkit~\citep{HTS2010} was used. Specifically, the implementation
from the EMIME project~\citep{Wester2010} was taken. Five-state,
left-to-right, no-skip HSMMs were used. The speech parameters which
were used for training the HSMMs were 39 order mel-cepstral
coefficients, log-F0 and 21-band aperiodicities, along with their
delta and delta-delta features, extracted every 5 ms.

For evaluating the phonological TTS, an intelligibility test
was conducted using semantically unpredictable sentences (SUSs). Two
sets of sentences were used in this test. 
Each set contained 14 unique SUSs.
The SUSs were taken from SIWIS database~\cite{Goldman_IS2016}. 
The length of the sentences varied from 6 to 8 words. Each set
consisted of 7 sentences synthesised by the phonological TTS, and
another 7 ones synthesised by the reference HTS system.

Twenty native English listeners, experts in the speech processing
field, participated in the listening test.
The listeners could listen to each synthesized sentence
only one or two times, and were asked to transcribe the audio. Eleven and
nine listeners respectively participated in the two sets of the
listening test.
Intelligibility score was calculated by:
\begin{equation}
\mbox{Intelligibility}=\frac{H-I}{N}\mbox{ x 100\%}
\end{equation}
where $H$ is the number of correctly transcribed words, $I$ is the
number of insertions, and $N$ is the total number of words in the
reference transcription.

The average intelligibility score of the phonological TTS was 71\%
in comparison to the HMM-based TTS where the listeners achieved the
intelligibility of 84\%. The phonological TTS thus achieved
intelligibility of 85\% of the state-of-the-art parametric TTS.

Proposed continuous speech synthesis from the phonological speech
representation is trained only from speech part of an audio-book
without the aligned text transcription, whereas conventional
parametric speech synthesis requires aligned phonetic labels for
training of the synthesis model. Hence, we consider this approach as
unsupervised TTS training.

{\rv \cite{Cernak16} have recently shown that major degradation of
  speech quality in speech synthesis  based on the phonological speech
  representation comes from the LPC re-synthesis. Therefore, other
  parametric vocoding is planned in our future work.}

\section{Conclusions}
\label{sec:conclusions}

We have proposed to use speech vocoding as a platform for laboratory
phonology. The proposal consists of a cascaded phonological analysis
and synthesis. The objective and subjective evaluations supported the
hypothesis that the most informative feature set -- with the best
coverage of the acoustic space (see confusion matrices of
context-independent vocoding in Sec.~\ref{sec:civ}) -- achieves the
best quality vocoded speech (see context-dependent vocoding in
Sec.~\ref{sec:cdv}), where the features are verified in both
directions, recognition and synthesis, simultaneously.

We have showed three application examples of our proposed
approach. First, we compared three systems of phonological
representations and concluded that eSPE achieves slightly better
results than the other two. Our results thus support other recent work
showing that eSPE is  suitable for phonological analysis, for speech
recognition and language identification tasks
\citep{Yu2012,Siniscalchi2012}. However, GP -- the most compact
phonological speech representation, performs in the
analysis/synthesis tasks comparably to the systems with higher number
of phonological features.

Second, we presented compositional phonological speech modelling,
where phonological atoms can generate arbitrary speech
sounds. Third, we explored phonological parametric TTS without
any front-end, trained from an unlabelled audiobook in an
unsupervised manner, and achieving intelligibility of 85\% of the
state-of-the-art parametric speech synthesis. 
This seems to be a promising approach for unsupervised and
multilingual text-to-speech systems.

In this work we have focused on the segmental evaluation of the
phonological systems. In the future, we plan to address this
limitation by incorporating supra-segmental features, and using the
proposed laboratory phonology platform for further experimentation
such as generation of speech stimuli for perception experiments.


We envision that the presented approach paves the way for researchers
in both fields to form meaningful hypotheses that are explicitly
testable using the concepts developed and exemplified in this
paper. Laboratory phonologists might test the compactness,
confusability, perceptual viability, and other applied concepts of
their theoretical models. This might be done in at least two
ways. Firstly, synthesis/recognition tests might follow hand in hand
with their analysis of data from human speech production and
perception, which would allow for accumulating much needed data on the
differences between human and machine performance. Secondly,
synthesis/recognition might be used for pre-testing before undergoing
experiments and analysis of human data, which is commonly time and
effort demanding. This framework can be used in speech processing
field as an evaluation platform for improving the performance of
current state-of-the-art applications, for example in multi-lingual
processing, using the concepts developed for the theoretical
phonological models. 



\section{Acknowledgements}
\label{sec:ack}

We would like to thank Dr. Phil~N.Garner from Idiap Research Institute,
for his valuable comments on this paper.

This work has been conducted with the support of the Swiss NSF under
grant CRSII2 141903: Spoken Interaction with Interpretation in Switzerland
(SIWIS) and grant 2/0197/15 by Scientific Granting Agency in Slovakia.



\section{References}
\label{sec:ref}

\bibliographystyle{elsarticle/elsarticle-harv}
\bibliography{refs,from_alex}

\appendix

\setcounter{table}{0}
\section{Mapping of the phonological features to CMUbet}
\label{sec:spefeatures}
Tables \ref{tab:GPatts}, \ref{tab:SPEatts} and
\ref{tab:eSPEatts} show the mapping of the phonological features to the
used phonemes in this work.

\begin{table} [t]
\footnotesize
  \caption{\label{tab:GPatts} {\it GP phonological features and
      their association to CMUbet phonemes used in this paper.}}
\vspace{2mm}
\centerline{
\begin{tabular}{ll|cccccccccccc}
\hline
\vt{CMUbet} & \vt{IPA} & A & I & U & E & S & h & H & N & a & i & u & \vt{silence} \\
\hline
iy & \ph{i}   & $-$ & $+$ & $-$ & $-$ & $-$ & $-$ & $-$ & $-$ & $-$ & $+$ & $-$ & $-$ \\
ih & \ph{I}   & $-$ & $+$ & $-$ & $+$ & $-$ & $-$ & $-$ & $-$ & $-$ & $-$ & $-$ & $-$ \\
uw & \ph{u}   & $-$ & $-$ & $+$ & $-$ & $-$ & $-$ & $-$ & $-$ & $-$ & $-$ & $+$ & $-$ \\
uh & \ph{U}   & $-$ & $-$ & $+$ & $+$ & $-$ & $-$ & $-$ & $-$ & $-$ & $-$ & $-$ & $-$ \\
ey & \ph{eI}  & $+$ & $+$ & $-$ & $-$ & $-$ & $-$ & $-$ & $-$ & $-$ & $+$ & $-$ & $-$ \\
ow & \ph{oU}  & $+$ & $-$ & $+$ & $-$ & $-$ & $-$ & $-$ & $-$ & $-$ & $-$ & $+$ & $-$ \\
oy & \ph{oI}  & $+$ & $-$ & $+$ & $-$ & $-$ & $-$ & $-$ & $-$ & $-$ & $+$ & $+$ & $-$ \\
ao & \ph{O}   & $+$ & $-$ & $+$ & $+$ & $-$ & $-$ & $-$ & $-$ & $-$ & $-$ & $+$ & $-$ \\
aa & \ph{A}   & $+$ & $-$ & $-$ & $-$ & $-$ & $-$ & $-$ & $-$ & $+$ & $-$ & $-$ & $-$ \\
ae & \ph{\ae} & $+$ & $+$ & $-$ & $-$ & $-$ & $-$ & $-$ & $-$ & $+$ & $-$ & $-$ & $-$ \\
ah & \ph{2}   & $+$ & $-$ & $-$ & $+$ & $-$ & $-$ & $-$ & $-$ & $-$ & $-$ & $-$ & $-$ \\
aw & \ph{aU}  & $+$ & $-$ & $+$ & $-$ & $-$ & $-$ & $-$ & $-$ & $-$ & $-$ & $+$ & $-$ \\
ay & \ph{aI}  & $+$ & $+$ & $-$ & $-$ & $-$ & $-$ & $-$ & $-$ & $-$ & $+$ & $-$ & $-$ \\
y  & \ph{j}   & $-$ & $+$ & $-$ & $-$ & $-$ & $-$ & $-$ & $-$ & $-$ & $-$ & $-$ & $-$ \\
w  & \ph{w}   & $-$ & $-$ & $+$ & $-$ & $-$ & $-$ & $-$ & $-$ & $-$ & $-$ & $-$ & $-$ \\
eh & \ph{e}   & $+$ & $+$ & $-$ & $+$ & $-$ & $-$ & $-$ & $-$ & $-$ & $+$ & $-$ & $-$ \\
er & \ph{3\textrhoticity}
              & $+$ & $-$ & $-$ & $+$ & $-$ & $-$ & $-$ & $-$ & $-$ & $-$ & $-$ & $-$ \\
r  & \ph{\*r} & $+$ & $-$ & $+$ & $+$ & $-$ & $-$ & $-$ & $-$ & $-$ & $-$ & $-$ & $-$ \\
l  & \ph{l}   & $-$ & $-$ & $-$ & $-$ & $+$ & $-$ & $-$ & $-$ & $-$ & $-$ & $-$ & $-$ \\
p  & \ph{p}   & $-$ & $-$ & $+$ & $-$ & $+$ & $+$ & $+$ & $-$ & $-$ & $-$ & $-$ & $-$ \\
b  & \ph{b}   & $-$ & $-$ & $+$ & $-$ & $+$ & $+$ & $-$ & $-$ & $-$ & $-$ & $-$ & $-$ \\
f  & \ph{f}   & $-$ & $-$ & $+$ & $-$ & $-$ & $+$ & $+$ & $-$ & $-$ & $-$ & $-$ & $-$ \\
v  & \ph{v}   & $-$ & $-$ & $+$ & $-$ & $-$ & $+$ & $-$ & $-$ & $-$ & $-$ & $-$ & $-$ \\
m  & \ph{m}   & $-$ & $-$ & $+$ & $-$ & $+$ & $-$ & $-$ & $+$ & $-$ & $-$ & $-$ & $-$ \\
t  & \ph{t}   & $+$ & $-$ & $-$ & $-$ & $+$ & $+$ & $+$ & $-$ & $-$ & $-$ & $-$ & $-$ \\
d  & \ph{d}   & $+$ & $-$ & $-$ & $-$ & $+$ & $+$ & $-$ & $-$ & $-$ & $-$ & $-$ & $-$ \\
th & \ph{T}   & $+$ & $-$ & $-$ & $-$ & $-$ & $+$ & $+$ & $-$ & $-$ & $-$ & $-$ & $-$ \\
dh & \ph{D}   & $+$ & $-$ & $-$ & $-$ & $-$ & $+$ & $-$ & $-$ & $-$ & $-$ & $-$ & $-$ \\
n  & \ph{n}   & $-$ & $-$ & $-$ & $-$ & $+$ & $-$ & $-$ & $+$ & $-$ & $-$ & $-$ & $-$ \\
s  & \ph{s}   & $-$ & $-$ & $-$ & $+$ & $-$ & $+$ & $+$ & $-$ & $-$ & $-$ & $-$ & $-$ \\
z  & \ph{z}   & $-$ & $-$ & $-$ & $+$ & $-$ & $+$ & $-$ & $-$ & $-$ & $-$ & $-$ & $-$ \\
ch & \ph{tS}  & $-$ & $+$ & $-$ & $-$ & $+$ & $-$ & $+$ & $-$ & $-$ & $-$ & $-$ & $-$ \\
jh & \ph{dZ}  & $-$ & $+$ & $-$ & $-$ & $+$ & $-$ & $-$ & $-$ & $-$ & $-$ & $-$ & $-$ \\
sh & \ph{S}   & $-$ & $+$ & $-$ & $-$ & $-$ & $+$ & $+$ & $-$ & $-$ & $-$ & $-$ & $-$ \\
zh & \ph{Z}   & $-$ & $+$ & $-$ & $-$ & $-$ & $+$ & $-$ & $-$ & $-$ & $-$ & $-$ & $-$ \\
k  & \ph{k}   & $-$ & $-$ & $-$ & $+$ & $+$ & $+$ & $+$ & $-$ & $-$ & $-$ & $-$ & $-$ \\
g  & \ph{g}   & $-$ & $-$ & $-$ & $+$ & $+$ & $+$ & $-$ & $-$ & $-$ & $-$ & $-$ & $-$ \\
ng & \ph{N}   & $-$ & $-$ & $-$ & $+$ & $+$ & $-$ & $-$ & $+$ & $-$ & $-$ & $-$ & $-$ \\
hh & \ph{h}   & $-$ & $-$ & $-$ & $-$ & $-$ & $+$ & $+$ & $-$ & $-$ & $-$ & $-$ & $-$ \\
\hline
\end{tabular}}
\end{table}

\begin{table} [t]
\footnotesize
  \caption{\label{tab:SPEatts} {\it SPE phonological features and
      their association to CMUbet phonemes used in this paper.}}
\vspace{2mm}
\centerline{
\begin{tabular}{ll|ccccccccccccccc}
\hline
\vt{CMUbet} & \vt{IPA} & \vt{vocalic} & \vt{consonantal} & \vt{high} &
\vt{back} & \vt{low} & \vt{anterior} & \vt{coronal} & \vt{round} &
\vt{rising} & \vt{tense} & \vt{voice} & \vt{continuant} & \vt{nasal} &
\vt{strident} & \vt{silence} \\
\hline
iy & \ph{i}   & $+$ & $-$ & $+$ & $-$ & $-$ & $-$ & $-$ & $-$ & $-$ & $+$ & $+$ & $+$ & $-$ & $-$ & $-$ \\
ih & \ph{I}   & $+$ & $-$ & $+$ & $-$ & $-$ & $-$ & $-$ & $-$ & $-$ & $-$ & $+$ & $+$ & $-$ & $-$ & $-$ \\
uw & \ph{u}   & $+$ & $-$ & $+$ & $+$ & $-$ & $-$ & $-$ & $+$ & $-$ & $+$ & $+$ & $+$ & $-$ & $-$ & $-$ \\
uh & \ph{U}   & $+$ & $-$ & $+$ & $+$ & $-$ & $-$ & $-$ & $+$ & $-$ & $-$ & $+$ & $+$ & $-$ & $-$ & $-$ \\
ey & \ph{eI}  & $+$ & $-$ & $-$ & $-$ & $-$ & $-$ & $-$ & $-$ & $+$ & $+$ & $+$ & $+$ & $-$ & $-$ & $-$ \\
ow & \ph{oU}  & $+$ & $-$ & $-$ & $+$ & $-$ & $-$ & $-$ & $+$ & $+$ & $+$ & $+$ & $+$ & $-$ & $-$ & $-$ \\
oy & \ph{oI}  & $+$ & $-$ & $-$ & $+$ & $-$ & $-$ & $-$ & $+$ & $+$ & $-$ & $+$ & $+$ & $-$ & $-$ & $-$ \\
ao & \ph{O}   & $+$ & $-$ & $-$ & $+$ & $-$ & $-$ & $-$ & $+$ & $-$ & $-$ & $+$ & $+$ & $-$ & $-$ & $-$ \\
aa & \ph{A}   & $+$ & $-$ & $-$ & $+$ & $+$ & $-$ & $-$ & $-$ & $-$ & $+$ & $+$ & $+$ & $-$ & $-$ & $-$ \\
ae & \ph{\ae} & $+$ & $-$ & $-$ & $-$ & $+$ & $-$ & $-$ & $-$ & $-$ & $-$ & $+$ & $+$ & $-$ & $-$ & $-$ \\
ah & \ph{2}   & $+$ & $-$ & $-$ & $+$ & $-$ & $-$ & $-$ & $-$ & $-$ & $-$ & $+$ & $+$ & $-$ & $-$ & $-$ \\
aw & \ph{aU}  & $+$ & $-$ & $-$ & $+$ & $+$ & $-$ & $-$ & $-$ & $+$ & $+$ & $+$ & $+$ & $-$ & $-$ & $-$ \\
ay & \ph{aI}  & $+$ & $-$ & $-$ & $-$ & $+$ & $-$ & $-$ & $-$ & $+$ & $+$ & $+$ & $+$ & $-$ & $-$ & $-$ \\
y  & \ph{j}   & $-$ & $-$ & $+$ & $-$ & $-$ & $-$ & $-$ & $-$ & $-$ & $-$ & $+$ & $+$ & $-$ & $-$ & $-$ \\
w  & \ph{w}   & $-$ & $-$ & $+$ & $+$ & $-$ & $-$ & $-$ & $+$ & $-$ & $-$ & $+$ & $+$ & $-$ & $-$ & $-$ \\
eh & \ph{e}   & $+$ & $-$ & $-$ & $-$ & $-$ & $-$ & $-$ & $-$ & $-$ & $-$ & $+$ & $+$ & $-$ & $-$ & $-$ \\
er & \ph{3\textrhoticity}
              & $+$ & $-$ & $-$ & $-$ & $-$ & $-$ & $-$ & $-$ & $-$ & $+$ & $+$ & $+$ & $-$ & $-$ & $-$ \\
r  & \ph{\*r} & $+$ & $+$ & $-$ & $-$ & $-$ & $-$ & $+$ & $-$ & $-$ & $-$ & $+$ & $+$ & $-$ & $-$ & $-$ \\
l  & \ph{l}   & $+$ & $+$ & $-$ & $-$ & $-$ & $+$ & $+$ & $-$ & $-$ & $-$ & $+$ & $+$ & $-$ & $-$ & $-$ \\
p  & \ph{p}   & $-$ & $+$ & $-$ & $-$ & $-$ & $+$ & $-$ & $-$ & $-$ & $-$ & $-$ & $-$ & $-$ & $-$ & $-$ \\
b  & \ph{b}   & $-$ & $+$ & $-$ & $-$ & $-$ & $+$ & $-$ & $-$ & $-$ & $-$ & $+$ & $-$ & $-$ & $-$ & $-$ \\
f  & \ph{f}   & $-$ & $+$ & $-$ & $-$ & $-$ & $+$ & $-$ & $-$ & $-$ & $-$ & $-$ & $+$ & $-$ & $+$ & $-$ \\
v  & \ph{v}   & $-$ & $+$ & $-$ & $-$ & $-$ & $+$ & $-$ & $-$ & $-$ & $-$ & $+$ & $+$ & $-$ & $+$ & $-$ \\
m  & \ph{m}   & $-$ & $+$ & $-$ & $-$ & $-$ & $+$ & $-$ & $-$ & $-$ & $-$ & $+$ & $-$ & $+$ & $-$ & $-$ \\
t  & \ph{t}   & $-$ & $+$ & $-$ & $-$ & $-$ & $+$ & $+$ & $-$ & $-$ & $-$ & $-$ & $-$ & $-$ & $-$ & $-$ \\
d  & \ph{d}   & $-$ & $+$ & $-$ & $-$ & $-$ & $+$ & $+$ & $-$ & $-$ & $-$ & $+$ & $-$ & $-$ & $-$ & $-$ \\
th & \ph{T}   & $-$ & $+$ & $-$ & $-$ & $-$ & $+$ & $+$ & $-$ & $-$ & $-$ & $-$ & $+$ & $-$ & $-$ & $-$ \\
dh & \ph{D}   & $-$ & $+$ & $-$ & $-$ & $-$ & $+$ & $+$ & $-$ & $-$ & $-$ & $+$ & $+$ & $-$ & $-$ & $-$ \\
n  & \ph{n}   & $-$ & $+$ & $-$ & $-$ & $-$ & $+$ & $+$ & $-$ & $-$ & $-$ & $+$ & $-$ & $+$ & $-$ & $-$ \\
s  & \ph{s}   & $-$ & $+$ & $-$ & $-$ & $-$ & $+$ & $+$ & $-$ & $-$ & $-$ & $-$ & $+$ & $-$ & $+$ & $-$ \\
z  & \ph{z}   & $-$ & $+$ & $-$ & $-$ & $-$ & $+$ & $+$ & $-$ & $-$ & $-$ & $+$ & $+$ & $-$ & $+$ & $-$ \\
ch & \ph{tS}  & $-$ & $+$ & $+$ & $-$ & $-$ & $-$ & $+$ & $-$ & $-$ & $-$ & $-$ & $-$ & $-$ & $+$ & $-$ \\
jh & \ph{dZ}  & $-$ & $+$ & $+$ & $-$ & $-$ & $-$ & $+$ & $-$ & $-$ & $-$ & $+$ & $-$ & $-$ & $+$ & $-$ \\
sh & \ph{S}   & $-$ & $+$ & $+$ & $-$ & $-$ & $-$ & $+$ & $-$ & $-$ & $-$ & $-$ & $+$ & $-$ & $+$ & $-$ \\
zh & \ph{Z}   & $-$ & $+$ & $+$ & $-$ & $-$ & $-$ & $+$ & $-$ & $-$ & $-$ & $+$ & $+$ & $-$ & $+$ & $-$ \\
k  & \ph{k}   & $-$ & $+$ & $+$ & $+$ & $-$ & $-$ & $-$ & $-$ & $-$ & $-$ & $-$ & $-$ & $-$ & $-$ & $-$ \\
g  & \ph{g}   & $-$ & $+$ & $+$ & $+$ & $-$ & $-$ & $-$ & $-$ & $-$ & $-$ & $+$ & $-$ & $-$ & $-$ & $-$ \\
ng & \ph{N}   & $-$ & $+$ & $+$ & $+$ & $-$ & $-$ & $-$ & $-$ & $-$ & $-$ & $+$ & $-$ & $+$ & $-$ & $-$ \\
hh & \ph{h}   & $-$ & $-$ & $-$ & $-$ & $+$ & $-$ & $-$ & $-$ & $-$ & $-$ & $-$ & $+$ & $-$ & $-$ & $-$ \\
\hline
\end{tabular}}
\end{table}

\begin{table} [t]
\footnotesize
  \caption{\label{tab:eSPEatts} {\it eSPE phonological features
      and their association to CMUbet phonemes used in this paper.}}
\vspace{2mm}
\centerline{
\begin{tabular}{ll|ccccccccccccccccccccc}
\hline
\vt{CMUbet} & \vt{IPA} & \vt{vowel} & \vt{fricative} & \vt{nasal} &
\vt{stop} & \vt{approxim.} & \vt{coronal} & \vt{high} & \vt{dental} &
\vt{glottal} & \vt{labial} & \vt{low} & \vt{mid} & \vt{retroflex} &
\vt{velar} & \vt{anterior} & \vt{back} & \vt{continuant} & \vt{round} &
\vt{tense} & \vt{voiced} & \vt{silence} \\
\hline
iy & \ph{i}   & $+$ & $-$ & $-$ & $-$ & $-$ & $-$ & $+$ & $-$ & $-$ & $-$ & $-$ & $-$ & $-$ & $-$ & $-$ & $-$ & $+$ & $-$ & $+$ & $+$ & $-$ \\
ih & \ph{I}   & $+$ & $-$ & $-$ & $-$ & $-$ & $-$ & $+$ & $-$ & $-$ & $-$ & $-$ & $-$ & $-$ & $-$ & $-$ & $-$ & $+$ & $-$ & $-$ & $+$ & $-$ \\
uw & \ph{u}   & $+$ & $-$ & $-$ & $-$ & $-$ & $-$ & $+$ & $-$ & $-$ & $-$ & $-$ & $-$ & $-$ & $-$ & $-$ & $+$ & $+$ & $+$ & $+$ & $+$ & $-$ \\
uh & \ph{U}   & $+$ & $-$ & $-$ & $-$ & $-$ & $-$ & $+$ & $-$ & $-$ & $-$ & $-$ & $-$ & $-$ & $-$ & $-$ & $+$ & $+$ & $+$ & $-$ & $+$ & $-$ \\
ey & \ph{eI}  & $+$ & $-$ & $-$ & $-$ & $-$ & $-$ & $-$ & $-$ & $-$ & $-$ & $-$ & $+$ & $-$ & $-$ & $-$ & $-$ & $+$ & $-$ & $+$ & $+$ & $-$ \\
ow & \ph{oU}  & $+$ & $-$ & $-$ & $-$ & $-$ & $-$ & $+$ & $-$ & $-$ & $-$ & $-$ & $+$ & $-$ & $-$ & $-$ & $+$ & $+$ & $+$ & $+$ & $+$ & $-$ \\
oy & \ph{oI}  & $+$ & $-$ & $-$ & $-$ & $-$ & $-$ & $-$ & $-$ & $-$ & $-$ & $-$ & $-$ & $-$ & $-$ & $-$ & $+$ & $+$ & $+$ & $-$ & $+$ & $-$ \\
ao & \ph{O}   & $+$ & $-$ & $-$ & $-$ & $-$ & $-$ & $-$ & $-$ & $-$ & $-$ & $-$ & $-$ & $-$ & $-$ & $-$ & $+$ & $+$ & $+$ & $+$ & $+$ & $-$ \\
aa & \ph{A}   & $+$ & $-$ & $-$ & $-$ & $-$ & $-$ & $-$ & $-$ & $-$ & $-$ & $+$ & $-$ & $-$ & $-$ & $-$ & $+$ & $+$ & $-$ & $+$ & $+$ & $-$ \\
ae & \ph{\ae} & $+$ & $-$ & $-$ & $-$ & $-$ & $-$ & $-$ & $-$ & $-$ & $-$ & $+$ & $-$ & $-$ & $-$ & $-$ & $-$ & $+$ & $-$ & $+$ & $+$ & $-$ \\
ah & \ph{2}   & $+$ & $-$ & $-$ & $-$ & $-$ & $-$ & $-$ & $-$ & $-$ & $-$ & $-$ & $+$ & $-$ & $-$ & $-$ & $+$ & $+$ & $-$ & $-$ & $+$ & $-$ \\
aw & \ph{aU}  & $+$ & $-$ & $-$ & $-$ & $-$ & $-$ & $-$ & $-$ & $-$ & $-$ & $+$ & $-$ & $-$ & $-$ & $-$ & $+$ & $+$ & $+$ & $+$ & $+$ & $-$ \\
ay & \ph{aI}  & $+$ & $-$ & $-$ & $-$ & $-$ & $-$ & $-$ & $-$ & $-$ & $-$ & $+$ & $-$ & $-$ & $-$ & $-$ & $+$ & $+$ & $-$ & $+$ & $+$ & $-$ \\
y  & \ph{j}   & $-$ & $-$ & $-$ & $-$ & $+$ & $-$ & $+$ & $-$ & $-$ & $-$ & $-$ & $-$ & $-$ & $-$ & $-$ & $-$ & $+$ & $+$ & $-$ & $+$ & $-$ \\
w  & \ph{w}   & $-$ & $-$ & $-$ & $-$ & $+$ & $-$ & $-$ & $-$ & $-$ & $+$ & $-$ & $-$ & $-$ & $-$ & $+$ & $-$ & $+$ & $+$ & $-$ & $+$ & $-$ \\
eh & \ph{e}   & $+$ & $-$ & $-$ & $-$ & $-$ & $-$ & $-$ & $-$ & $-$ & $-$ & $-$ & $+$ & $-$ & $-$ & $-$ & $-$ & $+$ & $-$ & $-$ & $+$ & $-$ \\
er & \ph{3\textrhoticity}
              & $+$ & $-$ & $-$ & $-$ & $-$ & $-$ & $-$ & $-$ & $-$ & $-$ & $-$ & $-$ & $+$ & $-$ & $-$ & $-$ & $+$ & $-$ & $-$ & $+$ & $-$ \\
r  & \ph{\*r} & $-$ & $-$ & $-$ & $-$ & $+$ & $-$ & $-$ & $-$ & $-$ & $-$ & $-$ & $-$ & $+$ & $-$ & $-$ & $-$ & $+$ & $+$ & $-$ & $+$ & $-$ \\
l  & \ph{l}   & $-$ & $-$ & $-$ & $-$ & $+$ & $+$ & $-$ & $-$ & $-$ & $-$ & $-$ & $-$ & $-$ & $-$ & $-$ & $-$ & $+$ & $-$ & $-$ & $+$ & $-$ \\
p  & \ph{p}   & $-$ & $-$ & $-$ & $+$ & $-$ & $-$ & $-$ & $-$ & $-$ & $+$ & $-$ & $-$ & $-$ & $-$ & $+$ & $-$ & $-$ & $-$ & $+$ & $-$ & $-$ \\
b  & \ph{b}   & $-$ & $-$ & $-$ & $+$ & $-$ & $-$ & $-$ & $-$ & $-$ & $+$ & $-$ & $-$ & $-$ & $-$ & $+$ & $-$ & $-$ & $-$ & $-$ & $+$ & $-$ \\
f  & \ph{f}   & $-$ & $+$ & $-$ & $-$ & $-$ & $-$ & $-$ & $-$ & $-$ & $+$ & $-$ & $-$ & $-$ & $-$ & $+$ & $-$ & $+$ & $-$ & $+$ & $-$ & $-$ \\
v  & \ph{v}   & $-$ & $+$ & $-$ & $-$ & $-$ & $-$ & $-$ & $-$ & $-$ & $+$ & $-$ & $-$ & $-$ & $-$ & $+$ & $-$ & $+$ & $+$ & $-$ & $+$ & $-$ \\
m  & \ph{m}   & $-$ & $-$ & $+$ & $-$ & $-$ & $-$ & $-$ & $-$ & $-$ & $+$ & $-$ & $-$ & $-$ & $-$ & $+$ & $-$ & $-$ & $-$ & $-$ & $+$ & $-$ \\
t  & \ph{t}   & $-$ & $-$ & $-$ & $+$ & $-$ & $+$ & $-$ & $-$ & $-$ & $-$ & $-$ & $-$ & $-$ & $-$ & $+$ & $-$ & $-$ & $-$ & $+$ & $-$ & $-$ \\
d  & \ph{d}   & $-$ & $-$ & $-$ & $+$ & $-$ & $+$ & $-$ & $-$ & $-$ & $-$ & $-$ & $-$ & $-$ & $-$ & $+$ & $-$ & $-$ & $-$ & $-$ & $+$ & $-$ \\
th & \ph{T}   & $-$ & $+$ & $-$ & $-$ & $-$ & $-$ & $-$ & $+$ & $-$ & $-$ & $-$ & $-$ & $-$ & $-$ & $+$ & $-$ & $+$ & $-$ & $+$ & $-$ & $-$ \\
dh & \ph{D}   & $-$ & $+$ & $-$ & $-$ & $-$ & $-$ & $-$ & $+$ & $-$ & $-$ & $-$ & $-$ & $-$ & $-$ & $+$ & $-$ & $+$ & $-$ & $-$ & $+$ & $-$ \\
n  & \ph{n}   & $-$ & $-$ & $+$ & $-$ & $-$ & $+$ & $-$ & $-$ & $-$ & $-$ & $-$ & $-$ & $-$ & $-$ & $+$ & $-$ & $-$ & $-$ & $-$ & $+$ & $-$ \\
s  & \ph{s}   & $-$ & $+$ & $-$ & $-$ & $-$ & $+$ & $-$ & $-$ & $-$ & $-$ & $-$ & $-$ & $-$ & $-$ & $+$ & $-$ & $+$ & $-$ & $+$ & $-$ & $-$ \\
z  & \ph{z}   & $-$ & $+$ & $-$ & $-$ & $-$ & $+$ & $-$ & $-$ & $-$ & $-$ & $-$ & $-$ & $-$ & $-$ & $+$ & $-$ & $+$ & $-$ & $-$ & $+$ & $-$ \\
ch & \ph{tS}  & $-$ & $+$ & $-$ & $-$ & $-$ & $-$ & $+$ & $-$ & $-$ & $-$ & $-$ & $-$ & $-$ & $-$ & $-$ & $-$ & $-$ & $-$ & $+$ & $-$ & $-$ \\
jh & \ph{dZ}  & $-$ & $+$ & $-$ & $-$ & $-$ & $-$ & $+$ & $-$ & $-$ & $-$ & $-$ & $-$ & $-$ & $-$ & $-$ & $-$ & $-$ & $-$ & $-$ & $+$ & $-$ \\
sh & \ph{S}   & $-$ & $+$ & $-$ & $-$ & $-$ & $-$ & $+$ & $-$ & $-$ & $-$ & $-$ & $-$ & $-$ & $-$ & $-$ & $-$ & $+$ & $-$ & $+$ & $-$ & $-$ \\
zh & \ph{Z}   & $-$ & $+$ & $-$ & $-$ & $-$ & $-$ & $-$ & $-$ & $-$ & $-$ & $-$ & $-$ & $-$ & $-$ & $-$ & $-$ & $-$ & $-$ & $-$ & $-$ & $-$ \\
k  & \ph{k}   & $-$ & $-$ & $-$ & $+$ & $-$ & $-$ & $+$ & $-$ & $-$ & $-$ & $-$ & $-$ & $-$ & $+$ & $-$ & $+$ & $-$ & $-$ & $+$ & $-$ & $-$ \\
g  & \ph{g}   & $-$ & $-$ & $-$ & $+$ & $-$ & $-$ & $+$ & $-$ & $-$ & $-$ & $-$ & $-$ & $-$ & $+$ & $-$ & $+$ & $-$ & $-$ & $-$ & $+$ & $-$ \\
ng & \ph{N}   & $-$ & $-$ & $+$ & $-$ & $-$ & $-$ & $+$ & $-$ & $-$ & $-$ & $-$ & $-$ & $-$ & $+$ & $-$ & $-$ & $-$ & $-$ & $-$ & $+$ & $-$ \\
hh & \ph{h}   & $-$ & $-$ & $-$ & $-$ & $-$ & $-$ & $-$ & $-$ & $+$ & $-$ & $-$ & $-$ & $-$ & $-$ & $-$ & $-$ & $-$ & $-$ & $-$ & $-$ & $-$ \\
\hline
\end{tabular}}
\end{table}

\setcounter{table}{0}
\section{Demonstrating speech synthesis samples}
\label{sec:samples}
Tables \ref{tab:SOPaudio} and \ref{tab:SOPcombaudio} contain
recordings demonstrating GP phonological atoms and their composition,
respectively, and Table \ref{tab:SPEaudio} contains recordings
demonstrating phonological speech synthesis.

\begin{table} [htb]
  \caption{\label{tab:SOPaudio} {\it Recordings demonstrating
      individual phonological atoms. The $z_n^k$ patterns were
      repeated to get phonological atoms $\vec{a}^k$ of two seconds
      long.}}
\vspace{2mm}
\centerline{
\begin{tabu}{|c|c|r|}
\hline
Phonolog. atom & Recording & Offline link\\
\hline \hline
$\vec{a}^A$ &
\textattachfile{audio/aa.wav}
{\includegraphics[height=1.8ex]{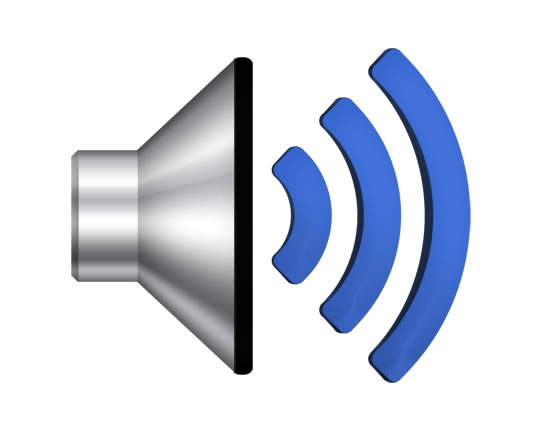}} (wav)
& \url{www.idiap.ch/paper/3120/A.wav}\\
\hline
$\vec{a}^a$ &
\textattachfile{audio/a.wav}
{\includegraphics[height=1.8ex]{figs/speaker-volume-button.jpg}} (wav)
& \url{www.idiap.ch/paper/3120/a.wav}\\
\hline
$\vec{a}^I$ &
\textattachfile{audio/ii.wav}
{\includegraphics[height=1.8ex]{figs/speaker-volume-button.jpg}} (wav)
& \url{www.idiap.ch/paper/3120/I.wav}\\
\hline
$\vec{a}^i$ &
\textattachfile{audio/i.wav}
{\includegraphics[height=1.8ex]{figs/speaker-volume-button.jpg}} (wav)
& \url{www.idiap.ch/paper/3120/i.wav}\\
\hline
$\vec{a}^U$ &
\textattachfile{audio/uu.wav}
{\includegraphics[height=1.8ex]{figs/speaker-volume-button.jpg}} (wav)
& \url{www.idiap.ch/paper/3120/U.wav}\\
\hline
$\vec{a}^u$ &
\textattachfile{audio/u.wav}
{\includegraphics[height=1.8ex]{figs/speaker-volume-button.jpg}} (wav)
& \url{www.idiap.ch/paper/3120/u.wav}\\
\hline
$\vec{a}^E$ &
\textattachfile{audio/E.wav}
{\includegraphics[height=1.8ex]{figs/speaker-volume-button.jpg}} (wav)
& \url{www.idiap.ch/paper/3120/E.wav}\\
\hline
$\vec{a}^H$ &
\textattachfile{audio/hh.wav}
{\includegraphics[height=1.8ex]{figs/speaker-volume-button.jpg}} (wav)
& \url{www.idiap.ch/paper/3120/H.wav}\\
\hline
$\vec{a}^h$ &
\textattachfile{audio/h.wav}
{\includegraphics[height=1.8ex]{figs/speaker-volume-button.jpg}} (wav)
& \url{www.idiap.ch/paper/3120/h.wav}\\
\hline
$\vec{a}^S$ &
\textattachfile{audio/S.wav}
{\includegraphics[height=1.8ex]{figs/speaker-volume-button.jpg}} (wav)
& \url{www.idiap.ch/paper/3120/S.wav}\\
\hline
$\vec{a}^N$ &
\textattachfile{audio/N.wav}
{\includegraphics[height=1.8ex]{figs/speaker-volume-button.jpg}} (wav)
& \url{www.idiap.ch/paper/3120/N.wav}\\
\hline
\end{tabu}}
\end{table}

\begin{table} [htb]
  \caption{\label{tab:SOPcombaudio} {\it Recordings demonstrating
      composition of GP phonological atoms, resulting into the
      synthesis of new sounds.}}
\vspace{2mm}
\centerline{
\begin{tabu}{|l|c|c|r|}
\hline
Rule & IPA & Composition & Offline link\\
\hline \hline
[A, I, U, E] & \ph{\oe} &
\textattachfile{audio/oe.wav}
{\includegraphics[height=1.8ex]{figs/speaker-volume-button.jpg}}
(wav) & \url{www.idiap.ch/paper/3120/oe.wav}\\
\hline
[I, U, E] & \ph{y} &
\textattachfile{audio/u.wav}
{\includegraphics[height=1.8ex]{figs/speaker-volume-button.jpg}}
(wav) & \url{www.idiap.ch/paper/3120/u.wav}\\
\hline
\end{tabu}}
\end{table}

\begin{table} [htb]
  \caption{\label{tab:SPEaudio} {\it Recordings demonstrating
      phonological speech synthesis.}}
\vspace{2mm}
\centerline{
\begin{tabu}{|l|c|c|r|}
\hline
Sentence & Scheme & Recording & Offline link\\
\hline \hline
\multirow{3}{*}{a0453} & GP &
\textattachfile{audio/gp-syn/a0453.wav}
{\includegraphics[height=1.8ex]{figs/speaker-volume-button.jpg}}
(wav) & \url{www.idiap.ch/paper/3120/a0453-GP.wav} \\
\cline{2-4}
 & SPE &
\textattachfile{audio/spe-syn/a0453.wav}
{\includegraphics[height=1.8ex]{figs/speaker-volume-button.jpg}}
(wav) & \url{www.idiap.ch/paper/3120/a0453-SPE.wav} \\
\cline{2-4}
 & eSPE &
\textattachfile{audio/pp-syn/a0453.wav}
{\includegraphics[height=1.8ex]{figs/speaker-volume-button.jpg}}
(wav) & \url{www.idiap.ch/paper/3120/a0453-eSPE.wav}\\
\hline
\multirow{3}{*}{a0457} & GP &
\textattachfile{audio/gp-syn/a0457.wav}
{\includegraphics[height=1.8ex]{figs/speaker-volume-button.jpg}}
(wav) & \url{www.idiap.ch/paper/3120/a0457-GP.wav} \\
\cline{2-4}
 & SPE &
\textattachfile{audio/spe-syn/a0457.wav}
{\includegraphics[height=1.8ex]{figs/speaker-volume-button.jpg}}
(wav) & \url{www.idiap.ch/paper/3120/a0457-SPE.wav}\\
\cline{2-4}
 & eSPE  &
\textattachfile{audio/pp-syn/a0457.wav}
{\includegraphics[height=1.8ex]{figs/speaker-volume-button.jpg}}
(wav) & \url{www.idiap.ch/paper/3120/a047-eSPE.wav}\\
\hline
\multirow{3}{*}{a0460} & GP &
\textattachfile{audio/gp-syn/a0460.wav}
{\includegraphics[height=1.8ex]{figs/speaker-volume-button.jpg}}
(wav) & \url{www.idiap.ch/paper/3120/a0460-GP.wav}\\
\cline{2-4}
 & SPE &
\textattachfile{audio/spe-syn/a0460.wav}
{\includegraphics[height=1.8ex]{figs/speaker-volume-button.jpg}}
(wav) & \url{www.idiap.ch/paper/3120/a0460-SPE.wav}\\
\cline{2-4}
 & eSPE &
\textattachfile{audio/pp-syn/a0460.wav}
{\includegraphics[height=1.8ex]{figs/speaker-volume-button.jpg}}
(wav) & \url{www.idiap.ch/paper/3120/a0460-eSPE.wav}\\
\hline
\end{tabu}}
\end{table}

\end{document}